\documentclass[runningheads]{llncs}

\usepackage{eccvabbrv}

\usepackage[utf8]{inputenc} 
\usepackage[T1]{fontenc}    
\usepackage{url}            
\usepackage{booktabs}       
\usepackage{amsfonts}       
\usepackage{nicefrac}       
\usepackage{microtype}      
\usepackage{etoolbox}
\usepackage{xcolor}         
\usepackage{graphicx}
\usepackage{enumitem}
\usepackage{changepage} 
\usepackage{placeins}
\usepackage{subcaption}
\usepackage[export]{adjustbox}
\usepackage{array}   
\usepackage{multirow}
\usepackage{algorithm}
\usepackage{algorithmic}
\usepackage{makecell}
\usepackage{svg}
\usepackage{xspace}
\usepackage{tabularx}
\usepackage{amsmath}
\usepackage{amssymb}

\usepackage{pifont}  
\newcommand{\cmark}{\textcolor{green}{\ding{51}}}
\newcommand{\xmark}{\textcolor{red}{\ding{55}}}
\newcommand{\br}[1]{\textcolor{red}{#1}} 
\newcommand{\bb}[1]{\textbf{\textcolor{blue}{#1}}}

 \newcommand{\li}[1]{#1}
 \newcommand{\jiaming}[1]{#1}

\newlength{\imgsz}
\newlength{\colw}
\usepackage[accsupp]{axessibility}  

\definecolor{eccvblue}{rgb}{0.21,0.49,0.74}
\definecolor{ctvidgreen}{RGB}{0,128,96}
\usepackage[breaklinks,colorlinks,allcolors=eccvblue]{hyperref}

\usepackage{orcidlink}

\newcommand{\equalcontrib}{\textsuperscript{*}}
\newcommand{\corrauth}{\textsuperscript{\ensuremath{\dagger}}}
\newcommand{\internnote}{\textsuperscript{\ensuremath{\ddagger}}}
\AtBeginEnvironment{thebibliography}{\linespread{0.98}\selectfont\setlength{\itemsep}{0pt}}
\newcommand{\ctvidlogo}{%
  \texorpdfstring{\raisebox{-0.2em}{\includegraphics[height=2.0em]{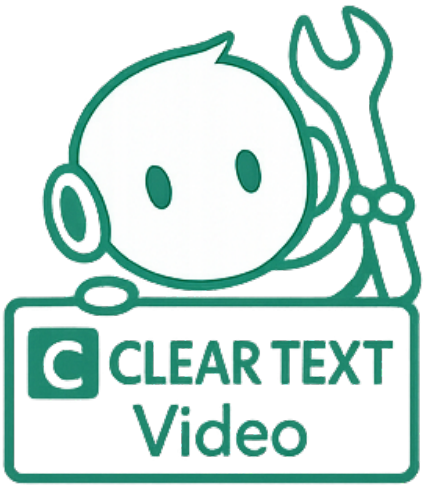}}}{ClearText-Video}%
  \xspace}

\begin{document}

\title{\texorpdfstring{%
\ctvidlogo {\color{ctvidgreen}ClearText-Video}: A Large-Scale Text-Centric Video Dataset Bridging Video Restoration and Scene-Text Enhancement\\[0.35em]
{\large\normalfont\href{https://jinlong17.github.io/CTVid-Bench/}{\textcolor{ctvidgreen}{\textbf{Project Page}}}}%
}{ClearText-Video: A Large-Scale Text-Centric Video Dataset Bridging Video Restoration and Scene-Text Enhancement}}

\titlerunning{ClearText-Video}

\author{ Jinlong Li\inst{1}\equalcontrib\corrauth\orcidlink{0000-0002-7784-8363} \and
Jiaming Ding\inst{1}\equalcontrib\orcidlink{0009-0004-0023-6248} \and
Dingfu Lu\inst{1,2}\internnote\orcidlink{0009-0000-1006-4372} \and
Malcolm Hsiu\inst{1,3}\internnote \and
Chuang Ke\inst{1}\orcidlink{0009-0004-3616-5509} \and
Kangning Yang\inst{1}\orcidlink{0000-0002-7106-0022} \and
Bochen Guan\inst{1}\orcidlink{0000-0003-1726-7214} \and
Lan Fu\inst{1}\orcidlink{0000-0003-2743-3116} \and
Jie Cai\inst{1}\orcidlink{0000-0001-6221-0319} \and
Huiming Sun\inst{1}\orcidlink{0000-0002-4329-7495} \and
Zibo Meng\inst{1}\orcidlink{0000-0001-7299-7290} }

\authorrunning{J.~Li et al.}

\institute{
OPPO US AI Center, USA
\and
University of Wisconsin--Madison, USA \quad \inst{3}University of California San Diego, USA\\
\email{\{jinlong.li1,jiaming.ding,chuang.ke3,kangning.yang,bochen.guan,lan.fu,}\\
\email{jie.cai,huiming.sun2,zibo.meng\}@oppo.com} 
}

\begingroup \renewcommand{\thefootnote}{} 
\footnotetext{ \hspace{-1.8em}\textsuperscript{*} Equal contribution. 
\quad\textsuperscript{\ensuremath{\dagger}} Corresponding author.\\
\hspace*{-1.8em}\textsuperscript{\ensuremath{\ddagger}} Work done during internships at OPPO US AI Center. } 
\endgroup

\maketitle

\begin{abstract}
Multimodal Large Language Models (MLLMs) have recently made strong progress in visual--linguistic understanding. However, their performance on text-centric video reasoning remains highly sensitive to input quality. Real-world user-provided videos often contain motion blur, compression artifacts, noise, and low-resolution text, which impair reliable text reading and downstream reasoning. Whether MLLMs can robustly read and reason about real-world scene text under diverse quality conditions remains a fundamental open question.
We introduce ClearText-Video (CTVid), a large-scale, scene-text-aware benchmark for studying text-centric video understanding under controlled quality variation. CTVid contains 4,639 real-world text-rich egocentric videos, 550K+ frames, 1.6M human-verified scene-text annotations, and 220K+ spatial/temporal question--answer pairs in Chinese and English. For each high-quality video, CTVid provides content-matched Degraded-Quality and Restored-Quality variants, supporting two task families: Text-Centric Video Restoration and Multi-Quality VideoQA.
We evaluate 18 representative restoration methods and 16 state-of-the-art MLLMs on CTVid. The results show that visual enhancement does not guarantee textual fidelity or downstream reasoning gains: blur is more damaging than low resolution, restored videos can alter the textual evidence used by MLLMs, and OCR-only pipelines remain far below direct multimodal reasoning. CTVid exposes the gap between video restoration and text-grounded understanding, providing a rigorous foundation for restoration-aware, quality-robust text-centric video systems.
\end{abstract}

\section{Introduction}

\begin{figure*}[ht]
    \centering
    \includegraphics[width=\textwidth]{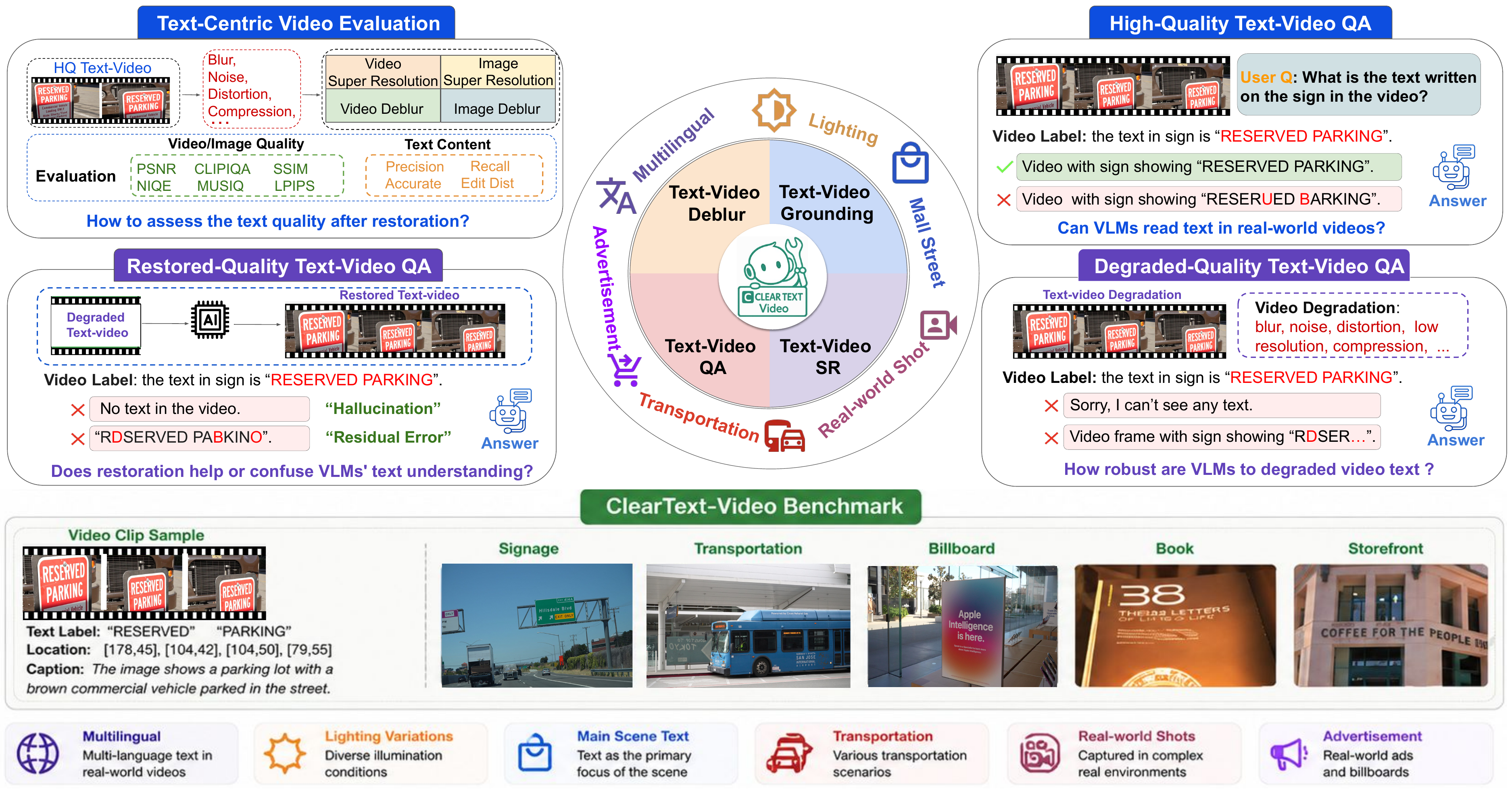}
    \caption{Overview of the \textbf{ClearText-Video (CTVid) Benchmark}. CTVid is built from bilingual (Zh/En) egocentric videos with scene-text boxes, transcripts, captions, and spatial/temporal QA pairs. Its key design is a quality-controlled video triplet: each content instance is evaluated as High-Quality (HQ), Degraded-Quality (DQ), and Restored-Quality (RQ), allowing direct analysis of how degradation and restoration change the textual evidence available to MLLMs. Unlike restoration-only datasets or single-quality TextVQA benchmarks, CTVid links Text-Centric Video Restoration (image SR, video SR, and deblurring) with Multi-Quality VideoQA, testing whether models can preserve, recover, and reason over the same scene text across changing video quality.}
    \label{fig:teaser}
\end{figure*}

Recent Multimodal Large Language Models (MLLMs)~\cite{yao2024minicpm,lin2024vila,fu2024vita,internvl2025v3,gemini2025v25} have made rapid progress in joint visual--linguistic reasoning and multimodal information processing. These capabilities now support a wide range of real-world applications, including autonomous driving, healthcare, and personal-assistant services~\cite{hu2024mplug,luo2024layoutllm,zhou2025egotextvqa,tom2023reading,jahagirdar2023watching}.
Amid this progress, Text-based Visual Question Answering (TextVQA) has emerged as a central yardstick for grounded multimodal intelligence, as it demands precise localization of visual evidence, accurate reading of scene text, and stepwise reasoning under real-world dynamics. VQA outcomes are sensitive to input quality and degrade when inputs exhibit poor resolution, compression artifacts, or blur. In real deployments, users often provide general-purpose systems (\textit{e.g.}, ChatGPT and Gemini) with suboptimal images or videos, which can reduce answer reliability.

Existing TextVQA benchmarks~\cite{zhou2025egotextvqa,tang2025mtvqa,zhou2024scene,jahagirdar2023watching} have substantially advanced the field. As shown in Table~\ref{tab:benchmark}, EgoTextVQA~\cite{zhou2025egotextvqa} introduces a text-aware VideoQA benchmark spanning outdoor driving and indoor housekeeping activities; MTVQA~\cite{tang2025mtvqa} provides an image-based QA benchmark for multilingual text scenarios; Real-CE~\cite{ma2023benchmark} establishes an image-based Chinese--English benchmark targeting text-image recovery;
NewsVideoQA~\cite{jahagirdar2023watching} poses questions about textual content in news videos, requiring models to read and reason to obtain answers; and ViTXT-GQA~\cite{zhou2024scene} enhances M4-ViteVQA~\cite{zhao2022towards} with spatiotemporal grounding annotations, enabling unified evaluation of answer grounding and QA.
Taken together, these benchmarks emphasize multilingual contexts and spatiotemporal text extraction, both of which are crucial for robust text-centric VideoQA.
However, existing datasets predominantly contain \textit{medium-to-low-resolution} images and videos. This limitation constrains systematic investigations of scene-text video quality and how resolution affects the text-centric capabilities of MLLMs.
A crucial question remains unanswered: \textit{Can MLLMs reliably read and reason over in-the-wild text under diverse quality conditions?}
To our knowledge, few studies explicitly examine multiple resolutions and quality levels, and current TextVQA benchmarks largely overlook resolution variation.

To address this question, we introduce ClearText-Video (CTVid), a large-scale, high-quality, scene-text-aware video question answering benchmark. ClearText-Video supports research on input quality and bilingual (Chinese/English) egocentric QA in real-world settings. We collect 4.6K+ real-world, text-centric high-quality videos, covering 550K+ frames and 1.6M annotated text instances. As illustrated in Fig.~\ref{fig:teaser}, ClearText-Video enables controlled analysis of input video quality through three regimes: High-Quality (HQ), Degraded-Quality (DQ), and Restored-Quality (RQ).
HQ features clean on-screen text and rich detail that favors MLLM encoding and reasoning; 
DQ is affected by blur and low resolution, which are typical in the wild and likely to impair text legibility and model performance; 
and RQ is produced by restoration models that aim to enhance video quality and recover unreadable text, with potential side effects on representation and reasoning.
Unlike prior restoration datasets, which focus on perceptual quality, and text-centric VideoQA datasets, which typically evaluate reasoning at a single quality level, CTVid combines controlled quality variation with multilingual scene-text challenges. It provides a unified benchmark that connects low-level restoration with high-level video understanding. We further evaluate 16 state-of-the-art MLLMs
and observe substantial headroom for improvement on CTVid, especially under quality variation.
Our main contributions are summarized as follows:
\begin{itemize}[topsep=0pt, itemsep=0pt, parsep=0pt]
    \item We introduce \underline{C}lear\underline{T}ext-\underline{Vid}eo (CTVid), a large-scale, high-quality, scene-text-aware video QA benchmark targeting input quality and bilingual (Zh/En) egocentric QA in real-world scenarios, with rich annotations (detection boxes, transcripts, and captions).
    \item We benchmark a range of MLLMs on CTVid and show persistent gaps in multilingual, quality-varying, text-rich settings, indicating significant room for advancement.
    \item Building on CTVid, we define two families of text-centric tasks: Text-Centric Video Restoration (including image super-resolution, video super-resolution, and video deblurring) and Multi-Quality Video Question Answering (covering spatial and temporal understanding). Together, these tasks establish a unified framework for evaluating textual fidelity and reasoning robustness under visual degradation, from low-level restoration to high-level reasoning.
\end{itemize}

\begin{table}[htbp]
\caption{Comparison of \textbf{ClearText-Video} with representative video datasets. ``Text in Video'' indicates whether textual content appears in video frames, and ``Content Language'' reports the languages of embedded text.}

\centering
\label{tab:benchmark}
\resizebox{0.99\textwidth}{!}{%
\begin{tabular}{@{}ccccccccc@{}}
\toprule
\textbf{Task}  & \textbf{Dataset}    & \textbf{Venue}    & \textbf{GT Resolution}    & \textbf{Caption} & \textbf{Text in Video} & \textbf{Content Language}  & \textbf{Videos/Img}  & \textbf{Video Source}      \\ \midrule
\multirow{6}{*}{Video Super-Resolution} & VideoLQ~\cite{chan2022investigating}      & CVPR 2022  & 640$\times$480          & \xmark   & \xmark        & \xmark             & 0.05K/4.9K   & Online \\
& SPMCS~\cite{tao2017detail}        & ICCV 2017  & 1920$\times$1080         & \xmark   & \xmark        & \xmark       & 0.03K/0.93K          & Manual           \\
& Vimeo-90k~\cite{xue2019video}    & IJCV 2019    & 448$\times$256           & \xmark       & \xmark             & \xmark     & 4.28K/0.6M            & Online  \\
& RealVSR~\cite{yang2021real}      & ICCV 2021  & 1024$\times$512  & \xmark   & \xmark        & \xmark            & 0.5K/2.5K     & Manual           \\
& YouHQ~\cite{zhou2024upscale}        & CVPR 2024  & 1920$\times$1080           & \xmark   & \xmark        & \xmark         & 37K/-       & Online \\
& MVSR4$\times$~\cite{wang2023benchmark}       & CVPRW 2023 & 1920$\times$1080        & \xmark   & \xmark        & \xmark           & 0.2K/2K     & Manual           \\ 
& Real-CE~\cite{ma2023benchmark}       & CVPR 2023 & 1920$\times$1080        & \xmark   & \xmark        & English, Chinese          & -/1.9K     & Manual  \\  
\midrule
\multirow{5}{*}{Video Deblurring} & GOPRO~\cite{nah2017deep}        & CVPR 2017   & 1280$\times$720            & \xmark   & \xmark        & \xmark        & -/3.2K        & Manual           \\
& BSD~\cite{zhong2023real}          & IJCV 2022   & 640$\times$480     & \xmark   & \xmark        & \xmark             & 0.08K/11K    & Manual           \\
& DVD~\cite{su2017deep}          & CVPR 2017   & 960$\times$540    & \xmark   & \xmark        & \xmark          & 0.07K/6.7K       & Manual           \\
& REVD~\cite{kim2024frequency}         & CVPR 2024   & 1024$\times$768     & \xmark   & \xmark        & \xmark             & 0.01K/6.3K    & Manual           \\
& REDS~\cite{nah2019ntire}         & NTIRE 2019  & 1280$\times$720         & \xmark   & \xmark        & \xmark            & 0.3K/30K    & Manual           \\ \midrule
\multirow{6}{*}{Visual Question Answering} & HD-VILA-100~\cite{xue2022advancing}  & CVPR 2022  & 1280$\times$720            & \cmark & \xmark        & \xmark           & -/3.3M       & YouTube          \\
& InternVid~\cite{wang2023internvid}    & ICLR 2024  & 1280$\times$720           & \cmark & \xmark        & \xmark            & -/7.1M       & YouTube          \\
& TVQA~\cite{lei2018tvqa}         & EMNLP 2018 & 1280$\times$720                  & \cmark & \xmark        & \xmark             & -/21.7K      & TV shows         \\
& RoadTextVQA~\cite{tom2023reading}  & ICDAR 2023  & 1280$\times$720               & \cmark & \xmark     & English         & -/3.2K   & YouTube          \\
& M4-ViteVQA~\cite{zhao2022towards}   & NeurIPS 2022 & 1280$\times$720            & \cmark     & \cmark           & English       & 7.6K/1.3M   & YouTube          \\
& NewsVideoQA~\cite{jahagirdar2023watching}  & WACV 2023  & 1280$\times$720       & \cmark & \cmark      & English          & 3K/0.9M  & YouTube          \\
&EgoTextVQA~\cite{zhou2025egotextvqa}  & CVPR 2025  & 960$\times$540       & \xmark & \cmark      & English          & 1.5K/-  & Manual          \\ 
& ViTXT-GQA~\cite{zhou2024scene}  & TMM 2025  & 1280$\times$720       & \cmark & \cmark      & English          & 7.6K/1.3M  & YouTube           \\ 
& MTVQA~\cite{tang2025mtvqa}  & ACL 2025  & 1280$\times$720                  & \cmark & \xmark      & 9 languages          & -/8.9K  & Manual          \\ 
& MME-VideoOCR~\cite{shi2025mme}  & arXiv 2025  & 1280$\times$720            & \cmark & \cmark      & English          & 1.4K/8.9K  & Manual    \\
\midrule
All &\textbf{ClearText-Video (Ours)}         & ECCV 2026  & 1920$\times$1080         & \cmark &\cmark & English, Chinese         & 4.6K/550K    & Manual           \\ \bottomrule
\end{tabular}%
}
\end{table}

\begin{figure}[htb]
    \centering
    \includegraphics[width=\textwidth]{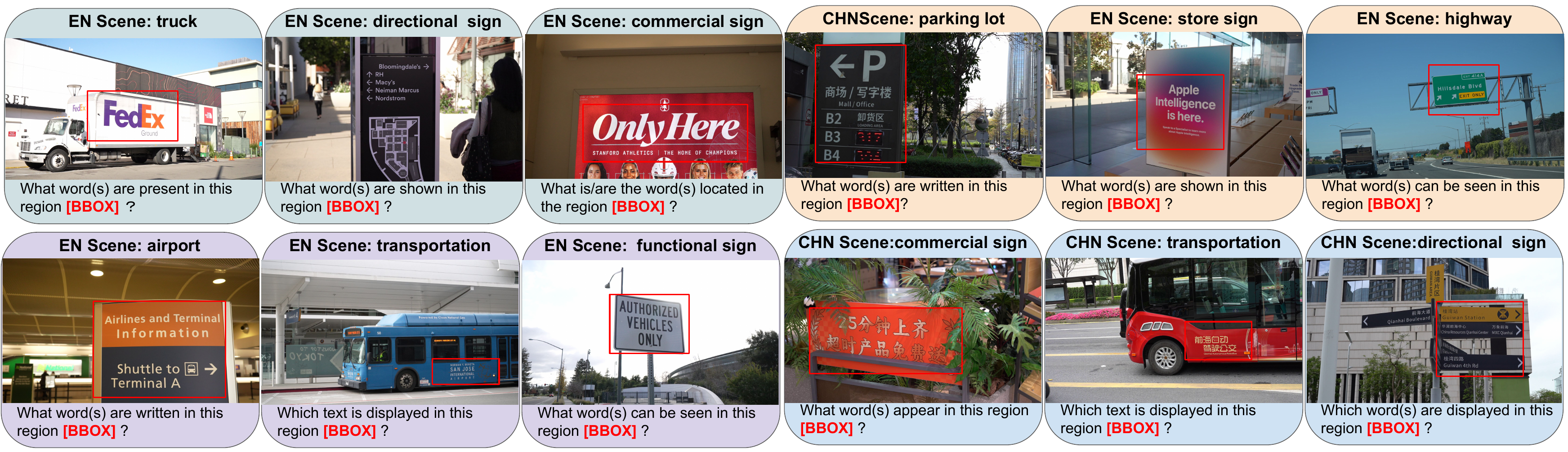}
    \caption{Example videos and their annotated questions from the ClearText-Video benchmark. Note: \br{[BBOX]} denotes bounding-box coordinates, visualized as red detection boxes in the images.}
    \label{fig:vis_videoQA}
\end{figure}

\section{Related Work}

\noindent \textbf{Real-world Video Restoration.}
Video restoration aims to recover high-quality video from low-quality footage degraded by imperfect capture, motion blur, or compression~\cite{wang2021real,yang2021real,wu2024seesr,liang2021swinir}. Building on the success of diffusion models in image restoration, recent methods extend them to video~\cite{zhou2024upscale,li2025diffvsr,yang2023mgldvsr,chen2025dove}. For example, Upscale-A-Video~\cite{zhou2024upscale} employs optical-flow-guided propagation, MGLD-VSR~\cite{yang2023mgldvsr} introduces motion-aware objectives, FlashVSR~\cite{zhuang2025flashvsr} proposes a one-step streaming framework for real-time restoration, and DiffVSR~\cite{li2025diffvsr} adopts staged optimization; DOVE~\cite{chen2025dove} leverages large-scale video-diffusion priors to create an efficient one-step diffusion model. Although these approaches markedly improve temporal consistency and perceptual quality, they provide limited treatment of semantics crucial for downstream tasks. Scene-text preservation remains underexplored, and generative restoration can hallucinate character-level details.

\noindent \textbf{Multimodal Large Language Models.}
MLLMs exhibit strong text-reading capabilities, making them well suited to text VQA tasks, including document understanding and text recognition~\cite{hu2024mplug,luo2024layoutllm}. Building on this foundation, recent MLLMs~\cite{yao2024minicpm,lin2024vila,fu2024vita,internvl2025v3,gemini2025v25} extend these capabilities to video, enabling the processing of dynamic visual information. Consequently, they can read text in static images and extract text signals from videos for more effective understanding.
However, performance still depends heavily on dataset diversity and scale. Furthermore, comprehensive, systematic evaluations of how input video quality affects text-centric performance remain limited, despite its central importance in real-world scenarios.

\noindent \textbf{Text-Aware VQA Benchmarks.}
In scene-text VQA, numerous datasets have been proposed. For example, TextVQA~\cite{singh2019towards}, ST-VQA~\cite{biten2019scene}, and ESTVQA~\cite{wang2020general} provide high-quality images with questions explicitly grounded in scene text; however, their image-based settings do not capture temporal reasoning in video. Recent work also broadens language coverage through multilingual resources, such as MTVQA~\cite{tang2025mtvqa} and EgoTextVQA~\cite{zhou2025egotextvqa}. On the video side, NewsVideoQA~\cite{jahagirdar2023watching} requires models to read and reason over on-screen text in news footage. ViTXT-GQA~\cite{zhou2024scene} extends M4-ViteVQA~\cite{zhao2022towards} with spatiotemporal grounding to jointly evaluate answer grounding and QA, and MME-VideoOCR~\cite{shi2025mme} covers a broad spectrum of video OCR scenarios to support deeper comprehension and reasoning. Meanwhile, RoadTextVQA~\cite{tom2023reading} provides text-rich videos, yet its questions remain largely simple and tightly focused on well-localized text, offering limited challenge for compositional reasoning and robustness.

\begin{figure*}[t]
    \centering
    \includegraphics[width=\linewidth]{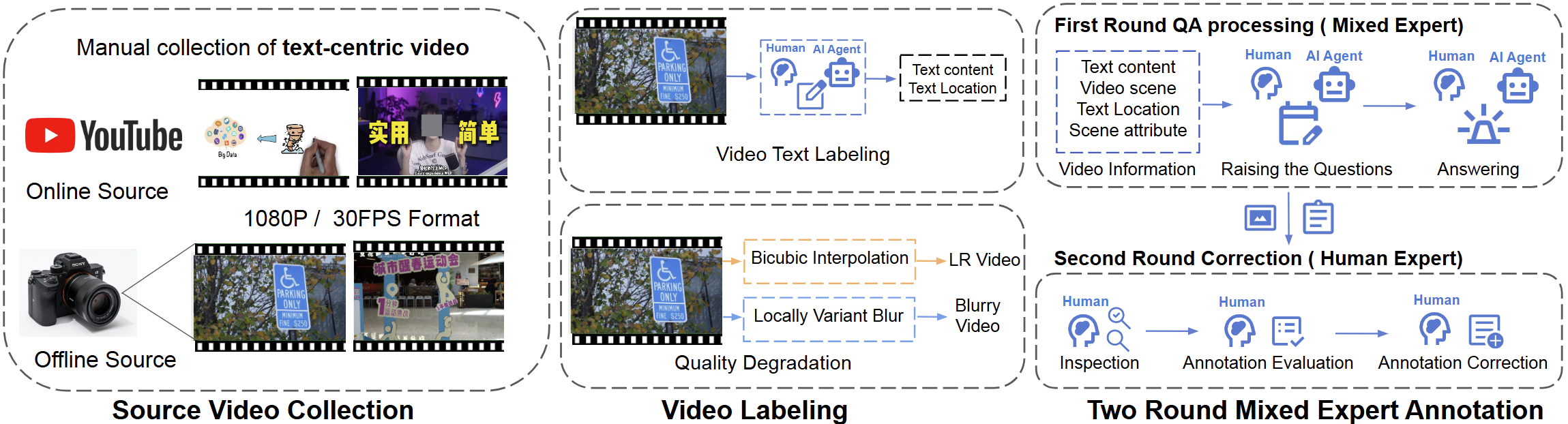}
    \caption{Overview of the annotation pipeline.}
    \label{fig:annotation_process}
\end{figure*}


\section{ClearText-Video Dataset}
Our benchmark targets multilingual, high-quality, text-rich video settings and is built from manually collected, text-centric footage rather than repurposing existing benchmarks. This section details the dataset's construction process, annotation schema, statistics, and the unified evaluation protocol.

\noindent \textbf{Video Collection.} 
We capture videos using a Sony A7R V with a 28--70 mm f/3.5--5.6 FE lens. The variable focal length supports recording at multiple zoom levels from diverse perspectives. Each 10-second raw video is center-cropped into a 4-second clip, from which we extract 120 frames for text annotation. To increase diversity, we source approximately 15\% of the videos from online text-centric content.

\noindent \textbf{Labeling and QA Generation.} 
The full annotation pipeline is illustrated in Fig.~\ref{fig:annotation_process}. Starting from more than 6.4K candidate videos, 16 annotators conduct two rounds of annotation and verification. After collecting and clipping the source videos, we annotate each HQ video frame with text detection boxes and transcripts. To reduce manual effort, we first use PaddleOCR~\cite{paddleocr} to generate coarse detection candidates, which annotators then refine with Labelme. The bounding boxes localize text regions, while the transcripts record the text appearing in those regions.
To create video-quality variants, we follow prior work~\cite{nah2019ntire} and apply bicubic downscaling ($\times 4$) to produce low-resolution videos. We also apply local motion blur to produce blurry videos. These synthetic degradations do not cover all real-world artifacts, but they create reproducible, content-matched variants that isolate resolution and blur effects while keeping the videos, text annotations, and QA pairs fixed.
We construct text-centric VQA pairs in two rounds, following annotation practices in~\cite{biten2019scene,zhou2025egotextvqa}. In the first round, we extract metadata for each text-rich video clip after labeling. The metadata records the text content, scene, text locations, and relevant scene attributes. We then prompt GPT-4o to propose a set of candidate question types (\textit{e.g.}, multiple choice, true/false, and fill-in-the-blank).
A rule-based pipeline then selects and instantiates candidate questions using the recorded text and location metadata, and derives the corresponding answers.
ClearText-Video focuses on two text-centric VideoQA settings: temporal and spatial understanding.
In the second round, to mitigate annotator and model biases and maintain high data quality, expert annotators visualize each clip and recheck text transcripts, text boxes, and their pairings. They verify answers and correct errors. They also screen for sensitive content, including privacy-related material. This round removes ambiguous questions, fixes inaccurate answers, and raises task difficulty where needed.

\begin{figure}[htb]
    \centering
    \includegraphics[width=1\textwidth]{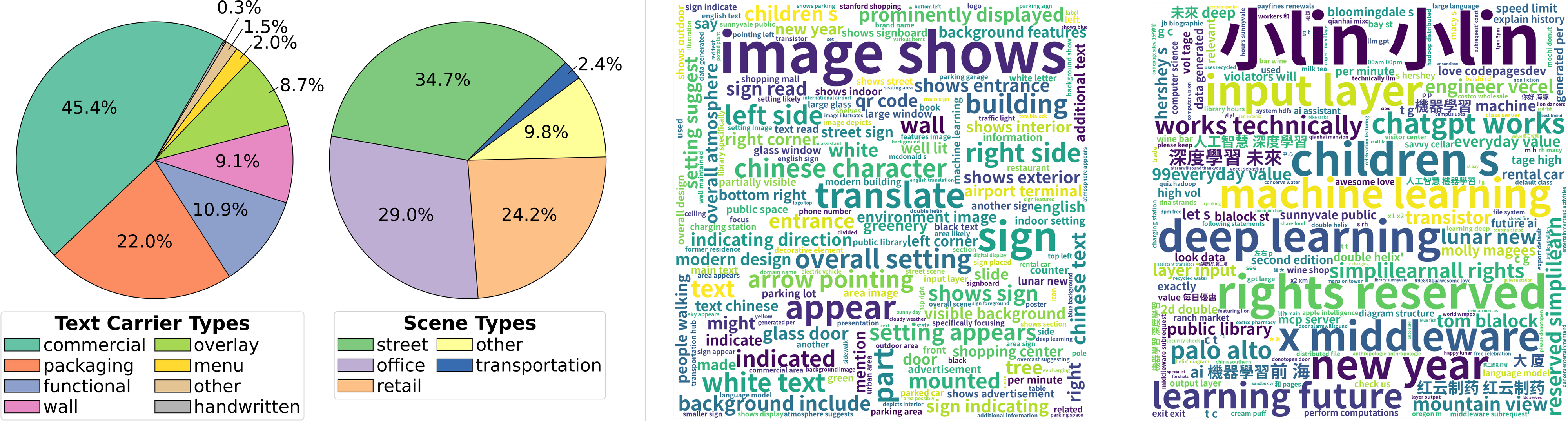}
    \caption{Statistics of the CTVid dataset. The left panels summarize text carrier types and scene categories, while the right panels show word-frequency visualizations for video captions and frame-level text annotations.}
    \label{fig:Distributions}
\end{figure}

\noindent \textbf{Dataset Statistics.}
The ClearText-Video benchmark contains 4,639 text-centric HQ videos, 550K+ frames, 1.6M annotated text instances, and 220K+ question--answer pairs. The dataset is split into 4,327 videos for training and 312 videos for testing. For each HQ video, we also provide low-resolution and blurry counterparts. The test set contains 74\% offline and 26\% online clips, balanced English/Chinese coverage within each source type, and a difficulty distribution of 46\% easy, 37\% medium, and 17\% hard questions. Fig.~\ref{fig:Distributions} summarizes the distributions of text carrier types and scene categories, together with word-frequency visualizations of captions and text annotations.

\noindent \textbf{Dataset Comparison.}
We compare ClearText-Video with related datasets in Table~\ref{tab:benchmark}. Among the datasets compared, restoration datasets in the first two task categories do not provide the annotations needed for text-aware restoration evaluation. ClearText-Video combines high-resolution video, explicit text regions, and frame-level captions, and offers higher resolution than the compared VideoQA datasets. Its central distinction is the quality-controlled setting: the same underlying videos support paired HQ/DQ/RQ evaluation, allowing us to test whether restoration preserves the textual evidence required for downstream reasoning rather than merely improving perceptual appearance.

\noindent \textbf{Data Release and Licensing}.
We will release train/test splits, human-verified boxes and transcripts, captions, temporal trajectories, spatial/temporal QA pairs, degradation scripts, evaluation code, prompts, and baseline outputs. Self-captured clips and annotations will be distributed under a research license. For online-source clips whose redistribution is restricted, we will release source IDs/URLs where allowed, clip ranges, annotations, and preprocessing scripts; raw videos will be redistributed only when permission or compatible licenses allow it.

\begin{table*}[t]
\centering
\caption{\textbf{Text-Centric Visual Super-Resolution Benchmark.} 
We report both general perceptual metrics and text-aware metrics. Text-aware metrics include precision (P$_{det}\%$), recall (R$_{det}\%$), and H-mean (F1$_{det}\%$) for detection, and accuracy (Acc$_{rec}\%$) and normalized edit distance (NED$_{rec}\%$) for recognition. The best and second-best results for each metric are highlighted in \textcolor{red}{\textbf{red}} and \textcolor{blue}{\textbf{blue}}, respectively.
}
\label{tab:benchmark_SR}
\resizebox{\textwidth}{!}{
\begin{tabular}{@{}cc|c|ccccccccc|cccccc@{}}
\toprule
& Input Type 
& Methods & PSNR$\uparrow$ & SSIM$\uparrow$ & LPIPS$\downarrow$ & DISTS$\downarrow$ & CLIPIQA$\uparrow$ & NIQE$\downarrow$ & MUSIQ$\uparrow$ & MANIQA$\uparrow$ & FID$\downarrow$ & P$_{det}$$\uparrow$ & R$_{det}$$\uparrow$ & F1$_{det}$$\uparrow$ & Acc$_{rec}$$\uparrow$ & NED$_{rec}$$\downarrow$ \\ \midrule
& \multirow{6}{*}{Image-based}
& Real-ESRGAN~\cite{wang2021real} & 27.85 & \bb{0.8748} & 0.1638 & 0.1032 & 0.4790 & 4.877 & 64.65 & 0.5718 & 35.00 & \br{89.06} & 57.78 & 70.09 & 34.82 & \bb{53.51} \\
& & SwinIR~\cite{liang2021swinir} & \br{28.18} & \br{0.8788} & \br{0.1613} & \bb{0.1012} & 0.5029 & 4.877 & 65.73 & 0.5705 & 32.15 & 88.78 & 58.47 & 70.50 & \bb{36.76} & 54.96 \\
& & SeeSR~\cite{wu2024seesr} & \bb{27.89} & 0.8529 & 0.1824 & 0.1161 & \br{0.6471} & 5.121 & \br{69.92} & \br{0.6196} & \br{27.66} & 87.27 & \bb{68.72} & \bb{76.89} & 35.08 & 54.95 \\
& & OSEDiff~\cite{wu2024one} & 25.67 & 0.8289 & 0.1946 & 0.1151 & 0.6219 & \bb{4.852} & 69.11 & \bb{0.6168} & 32.67 & 87.59 & 66.35 & 75.50 & 34.25 & 54.06 \\
& & S3Diff~\cite{zhang2024degradation} & 26.05 & 0.8033 & \bb{0.1617} & \br{0.09611} & 0.5764 & \br{4.517} & 64.52 & 0.5799 & \bb{29.16} & 86.72 & \br{69.52} & \br{77.17} & \br{38.42} & 58.03 \\
& & AdcSR~\cite{chen2025adversarial} & 25.84 & 0.8239 & 0.2024 & 0.124 & \bb{0.6448} & 4.941 & \bb{69.82} & 0.6083 & 37.52 & \bb{88.84} & 57.72 & 69.98 & 32.02 & \br{51.26} \\

\midrule
& \multirow{6}{*}{Video-based} 
& RealBasicVSR~\cite{chan2022investigating} & 28.26 & \bb{0.8924} & 0.1639 & 0.1126 & 0.4586 & \br{4.413} & 65.40 & \bb{0.6042} & 29.91 & 87.78 & 57.15 & 69.23 & \bb{41.97} & 59.30 \\
& & BasicVSR++~\cite{chan2022basicvsrpp} & 22.26 & 0.6214 & 0.4012 & 0.2429 & \br{0.5808} & 5.901 & \br{67.49} & \br{0.6358} & \br{18.02} & \br{91.11} & 54.50 & 68.20 & 35.21 & 55.34 \\
& & RealViFormer~\cite{zhang2024realviformer} & 27.95 & 0.8821 & \bb{0.1577} & 0.1198 & 0.4032 & 5.049 & 60.33 & 0.5521 & 32.56 & 88.74 & 48.98 & 63.12 & 29.89 & \br{46.37} \\
& & MGLD-VSR~\cite{yang2023mgldvsr} & \bb{28.98} & 0.8813 & 0.1596 & \bb{0.1084} & 0.4192 & \bb{4.510} & 63.4 & 0.5819 & \bb{19.66} & 90.21 & \bb{61.23} & \bb{72.95} & 39.66 & 58.66 \\
& & Upscale-A-Video~\cite{zhou2024upscale} & 27.64 & 0.8481 & 0.1875 & 0.1167 & \bb{0.5505} & 4.693 & \bb{66.96} & 0.5868 & 26.58 & \bb{90.62} & 58.86 & 71.36 & 33.40 & \bb{52.96} \\
& & DOVE~\cite{chen2025dove} & \br{30.24} & \br{0.9047} & \br{0.1267} & \br{0.08996} & 0.4185 & 5.210 & 63.91 & 0.5479 & 25.22 & 90.5 & \br{63.36} & \br{74.53} & \br{44.72} & 63.14 \\

\bottomrule
\end{tabular}
}
\end{table*}

\begin{table*}[t]
\centering
\caption{\textbf{Text-Centric Video Deblurring Benchmark.} We use the same metrics as the Super-Resolution Benchmark above. Best and second-best results are highlighted in \textcolor{red}{\textbf{red}} and \textcolor{blue}{\textbf{blue}}, respectively.}
\label{tab:benchmark_deblur}
\resizebox{\textwidth}{!}{
    \begin{tabular}{@{}cc|ccccccccc|ccccc@{}}
    \toprule
    & Methods & PSNR$\uparrow$ & SSIM$\uparrow$ & LPIPS$\downarrow$ & DISTS$\downarrow$ & CLIPIQA$\uparrow$ & NIQE$\downarrow$ & MUSIQ$\uparrow$ & MANIQA$\uparrow$ & FID$\downarrow$ & P$_{det}$$\uparrow$ & R$_{det}$$\uparrow$ & F1$_{det}$$\uparrow$ & Acc$_{rec}$$\uparrow$ & NED$_{rec}$$\downarrow$ \\ \midrule
    & MIMO-UNet+~\cite{cho2021rethinking} & 29.26 & 0.8657 & 0.2053 & 0.1393 & 0.3324 & 6.483 & 46.02 & 0.4760 & 47.15 & 88.58 & 65.02 & 74.99 & 51.75 & 69.54 \\
    & NAFNet~\cite{chen2022simple} & 28.21 & 0.8716 & 0.2461 & 0.1749 & \bb{0.3487} & 7.614 & 48.35 & 0.4253 & 48.65 & \bb{90.81} & 66.37 & 76.69 & 50.77 & 68.30 \\
    & Restormer~\cite{zamir2022restormer} & \br{31.38} & \br{0.8863} & 0.1904 & 0.1349 & 0.3404 & 6.552 & 47.31 & 0.4864 & 41.50 & 88.28 & 67.08 & 76.23 & 54.99 & 71.67 \\
    & Stripformer(GoPro)~\cite{tsai2022stripformer} & \bb{31.13} & \bb{0.8832} & \br{0.1754} & 0.1191 & 0.3348 & \bb{6.386} & 48.65 & \br{0.4955} & 38.78 & 88.42 & 66.69 & 76.03 & 55.05 & 72.26 \\
    & Stripformer(RealBlur-J)~\cite{tsai2022stripformer} & 27.04 & 0.8655 & \bb{0.1837} & \bb{0.1182} & \br{0.3657} & \br{6.240} & \br{54.44} & \bb{0.4922} & \br{32.36} & 89.18 & \br{69.99} & \br{78.43} & \br{58.93} & 76.51 \\
    & Stripformer(RealBlur-R)~\cite{tsai2022stripformer} & 27.36 & 0.8667 & 0.1905 & \br{0.1166} & 0.3396 & 6.490 & \bb{50.41} & 0.4668 & \bb{33.46} & 89.45 & \bb{69.71} & \bb{78.36} & \bb{58.57} & 75.71 \\
    & RVRT(DVD)~\cite{liang2022rvrt} & 28.83 & 0.8489 & 0.2671 & 0.1843 & 0.2902 & 7.158 & 34.37 & 0.4138 & 58.32 & 90.61 & 61.32 & 73.14 & 44.19 & \bb{60.61} \\
    & RVRT(GoPro)~\cite{liang2022rvrt} & 28.77 & 0.8553 & 0.2380 & 0.1622 & 0.2959 & 6.879 & 38.92 & 0.4407 & 52.91 & 89.63 & 61.25 & 72.77 & 46.23 & 63.31 \\
    & ShiftNet(DVD)~\cite{Li_2023_CVPR} & 28.47 & 0.8445 & 0.2771 & 0.1913 & 0.3077 & 7.334 & 33.52 & 0.4065 & 58.76 & \br{91.05} & 61.15 & 73.16 & 43.10 & \br{59.56} \\
    & ShiftNet(GoPro)~\cite{Li_2023_CVPR} & 28.20 & 0.8434 & 0.2570 & 0.1763 & 0.3004 & 6.965 & 35.69 & 0.4244 & 56.64 & 90.47 & 60.54 & 72.54 & 43.88 & 61.40 \\
    \bottomrule
    \end{tabular}
}
\end{table*}

\begin{figure*}[htb]
    \centering
    \includegraphics[width=\linewidth]{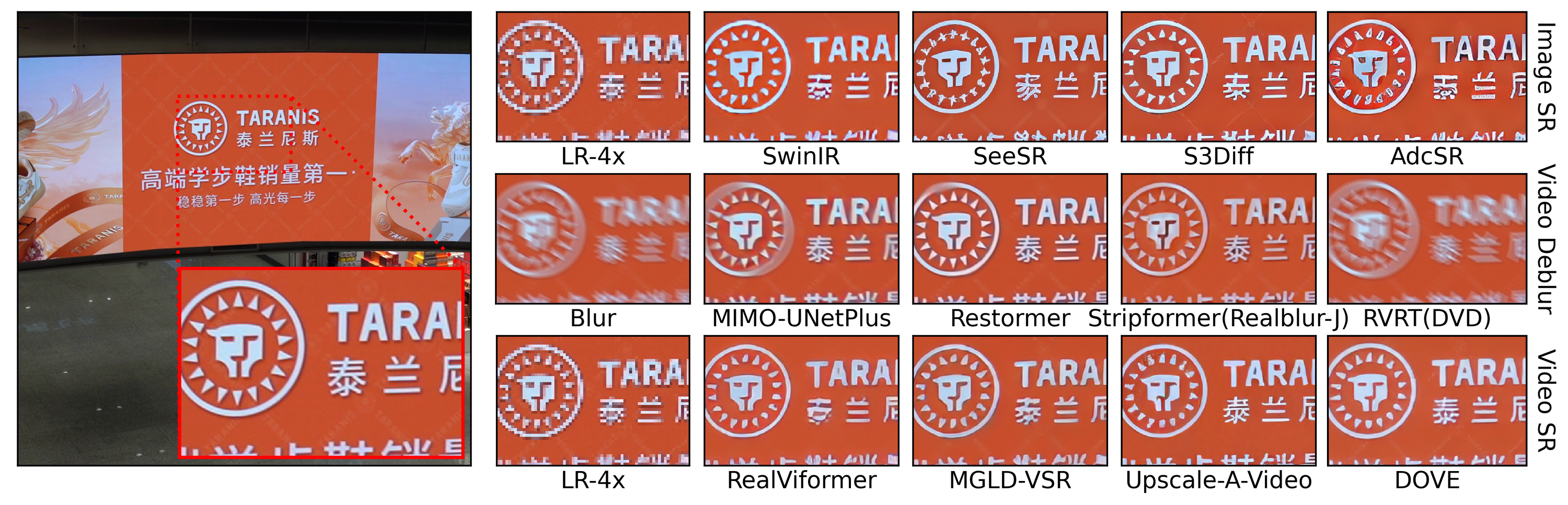}
    \includegraphics[width=\linewidth]
    {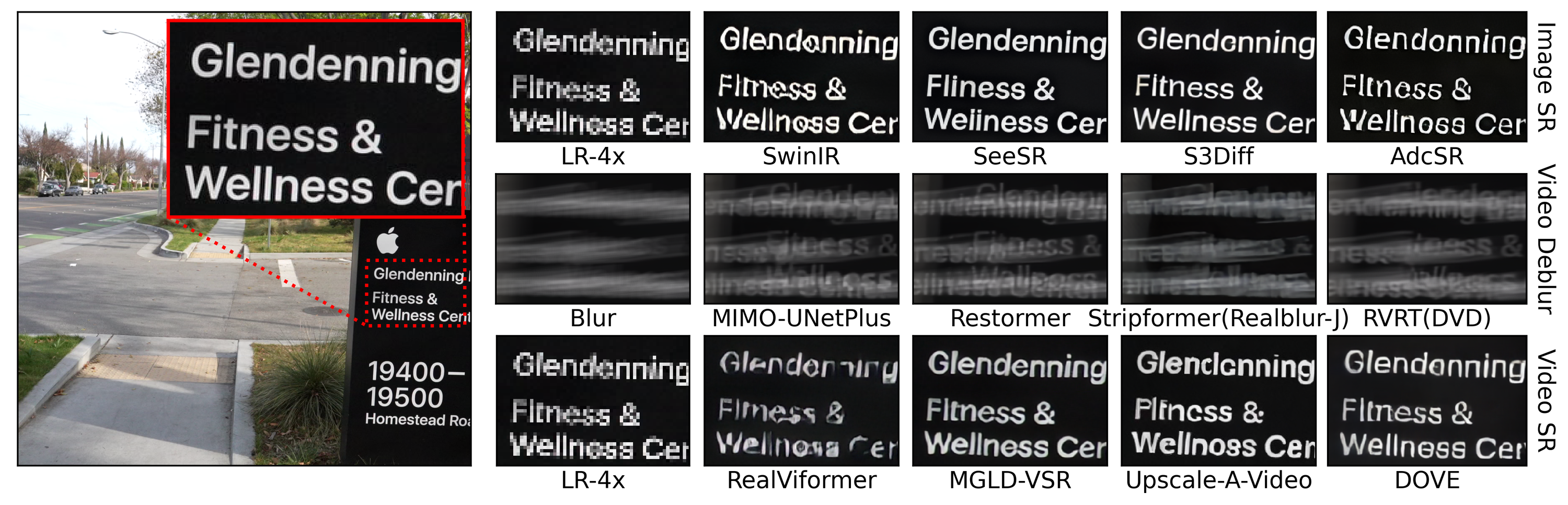}
    \caption{Qualitative results on the Text-Centric Video Restoration benchmarks on the CTVid test set. Results are shown for both English and Chinese text instances. The examples include video super-resolution and deblurring outputs, highlighting differences in text fidelity and legibility across methods.}
    \label{fig:Vis_results}
\end{figure*}

\section{ClearText-Video Benchmark}\label{Tasks}

Our ClearText-Video benchmark is designed to support text-centric video tasks across both low-level restoration and high-level understanding. It provides paired, temporally aligned videos for restoration tasks, specifically image and video super-resolution as well as video deblurring.
Building on this foundation, we introduce a multi-quality VideoQA benchmark for text-centric visual understanding, constructed from triplets of High-Quality (HQ), Degraded-Quality (DQ), and Restored-Quality (RQ) videos. This design enables rigorous evaluation of MLLMs under varying input quality and supports analyses of their text-semantic understanding across both spatial and temporal domains.

\subsection{Text-Centric Video Restoration}
\noindent \textbf{Text-Centric Image SR task.}
Text-centric image super-resolution aims to reconstruct high-resolution frames from 4$\times$ bicubic downsampled inputs~\cite{nah2019ntire}. Each CTVid frame includes a caption, enabling research on semantic-aware or language-guided super-resolution, where text aids the restoration of degraded visuals.

\noindent \textbf{Text-Centric Video SR task.}
Text-centric video super-resolution reconstructs high-resolution videos from low-resolution inputs. If semantic information is needed, the caption of the middle frame serves as the video's representative caption.

\noindent \textbf{Text-Centric Video Deblurring task.}
The Text-Centric Video Deblurring task seeks to recover sharp images from synthetically blurred inputs. Blur is synthesized by applying 16$\times$ frame interpolation with RIFE~\cite{huang2022rife} to convert 30 fps clips to 480 fps, followed by temporal fusion of every 65 frames to emulate camera or object motion.

\subsection{Text-Centric Multi-Quality VideoQA}

\noindent \textbf{Text-Centric VideoQA for Spatial Understanding task.}
This task focuses on understanding scene text within its spatial context in video frames. For each frame, we generate spatially grounded questions in multiple-choice, true/false, and fill-in-the-blank formats, based on text that appears inside a specific bounding-box region. A rule-based spatial partitioning method identifies meaningful regions by analyzing the layout of detected text and selecting partition boundaries to form bounding boxes that isolate specific text groups. GPT-4o then formulates natural-language questions using these regions, while a validation module ensures that the questions are spatially correct, well-structured, and answerable.
We then manually verify that the questions cannot be answered by text matching alone and require the specified spatial relation.

\noindent \textbf{Text-Centric VideoQA for Temporal Understanding task.}
This task evaluates a model's ability to understand how scene text moves and changes over time within a video. Given 120-frame clips, questions require reasoning about temporal visibility patterns, motion behavior, spatial transitions, and size variations of text instances. All questions are generated deterministically through rule-based analysis of text trajectories. For each text instance, we track frame-by-frame properties such as position, area, speed, and acceleration. We then use these temporal features to construct questions across five categories: visibility, spatial localization over time, motion and trajectory shape, size and scale changes, and boundary interactions. After rule-based generation, annotators screen and refine the outputs so that only meaningful, high-quality questions are retained. This hybrid pipeline supports consistency and reproducibility while reducing the risk of LLM hallucinations.

\subsection{Evaluation Protocol.} 
\noindent \textit{A) Evaluation on visual restoration.} 
We evaluate restoration effectiveness with both image-quality and text-centric measures. For image quality assessment, we report both full-reference and no-reference metrics. For full-reference IQA, we use PSNR, SSIM~\cite{SSIM}, LPIPS~\cite{zhang2018unreasonable}, DISTS~\cite{DISTS}, and FID~\cite{heusel2017gans}; for no-reference IQA, we use NIQE~\cite{niqe}, MANIQA~\cite{maniqa}, MUSIQ~\cite{musiq}, and CLIPIQA~\cite{clipiqa}. To measure textual fidelity in restored regions, we run a fixed PaddleOCR~\cite{paddleocr} pipeline and report detection and recognition metrics as indirect proxies.
We adopt precision (P$_{det}$), recall (R$_{det}$), and H-mean (F1$_{det}$) to measure text detection performance, while text recognition accuracy (Acc$_{rec}$) and normalized edit distance (NED$_{rec}$) are used to measure text recognition performance.

\noindent \textit{B) Evaluation on visual question answering.}
We evaluate VideoQA with four metrics: accuracy (Acc), uncertainty-aware accuracy (UAcc), overconfidence ratio (OC), and abstention rate (Abs). UAcc and OC follow the definitions in~\cite{xia2024cares}, assessing reliability by coupling correctness with confidence. Abs measures the fraction of queries on which the model withholds an answer~\cite{autotrust}.

\begin{table*}[htbp]
\caption{\textbf{Text-Centric VideoQA for Spatial Understanding.} 
We evaluate accuracy (Acc), uncertainty-aware accuracy (UAcc), overconfidence ratio (OC), and abstention rate (Abs) across six input-quality conditions: HQ (High-Quality), DQ-Low\_res~\cite{nah2019ntire} (low-resolution degradation), DQ-Blur~\cite{nah2019ntire} (blur degradation), RQ-DOVE~\cite{chen2025dove} (video-based SR), RQ-MIMO~\cite{cho2021rethinking} (blur restoration), and RQ-S3DIFF~\cite{zhang2024degradation} (image-based SR).
``Qwen2.5-VL-7B-SFT'' denotes Qwen2.5-VL-7B instruction-tuned on the ClearText-Video training set.
For Acc, OC, and UAcc, the best and second-best results in each column are highlighted in \textcolor{red}{\textbf{red}} and \textcolor{blue}{\textbf{blue}}, respectively. All reported values are percentages (\%).
}
\label{tab:all-imageQA_results}
\resizebox{\textwidth}{!}
{%
\begin{tabular}{@{}l|cccc|cccc|cccc|cccc|cccc|cccc@{}}
\toprule \midrule &
  \multicolumn{4}{c|}{HQ} &
  \multicolumn{4}{c|}{DQ-Low\_res~\cite{nah2019ntire}} &
  \multicolumn{4}{c|}{DQ-Blur~\cite{nah2019ntire}} &
  \multicolumn{4}{c|}{RQ-DOVE~\cite{chen2025dove}} &
  \multicolumn{4}{c|}{RQ-MIMO~\cite{cho2021rethinking}} &
  \multicolumn{4}{c}{RQ-S3DIFF~\cite{zhang2024degradation}} \\
Model &
  Acc$\uparrow$ & OC$\downarrow$ & UAcc$\uparrow$ & Abs &
  Acc$\uparrow$ & OC$\downarrow$ & UAcc$\uparrow$ & Abs &
  Acc$\uparrow$ & OC$\downarrow$ & UAcc$\uparrow$ & Abs &
  Acc$\uparrow$ & OC$\downarrow$ & UAcc$\uparrow$ & Abs &
  Acc$\uparrow$ & OC$\downarrow$ & UAcc$\uparrow$ & Abs &
  Acc$\uparrow$ & OC$\downarrow$ & UAcc$\uparrow$ & Abs \\ \midrule 
GPT-5.4~\cite{openai2026gpt54} & 60.00 & 15.00 & 55.00 & 0.00 & 53.33 & 10.00 & 58.33 & 6.70 & 56.67 & 8.33 & 60.00 & 18.30 & 48.33 & 15.00 & 55.00 & 3.30 & 55.00 & \bb{10.00} & 58.33 & 6.70 & 53.33 & 13.33 & 53.33 & 0.00 \\
GPT-5.4-mini~\cite{openai2026gpt54mini} & 46.67 & \bb{6.67} & \bb{65.00} & 0.00 & 48.33 & 18.33 & 51.67 & 0.00 & 43.33 & \bb{5.00} & 63.33 & 3.30 & 46.67 & \bb{3.33} & 66.67 & 0.00 & 41.67 & 11.67 & 56.67 & 0.00 & 45.00 & \bb{8.33} & 56.67 & 0.00 \\
Claude-Sonnet-4.6~\cite{anthropic2026claudesonnet} & \bb{70.00} & 21.67 & \br{66.67} & 0.00 & 63.33 & 20.00 & \bb{70.00} & 0.00 & \br{60.00} & 20.00 & \bb{70.00} & 13.30 & \bb{65.00} & 18.33 & \br{70.00} & 0.00 & \br{70.00} & 20.00 & \br{70.00} & 3.30 & \bb{61.67} & 16.67 & \br{71.67} & 0.00 \\
Gemini-2.5-pro~\cite{gemini2025v25} & \br{71.67} & 16.67 & 65.00 & 0.00 & \br{65.00} & \bb{8.33} & \br{71.67} & 1.70 & \bb{60.00} & 13.33 & 66.67 & 23.30 & 65.00 & 13.33 & \bb{68.33} & 1.70 & \bb{63.33} & 20.00 & 56.67 & 10.00 & \br{70.00} & 11.67 & \bb{65.00} & 0.00 \\
Gemini-2.5-flash~\cite{gemini2025v25} & 60.00 & 21.67 & 61.67 & 1.70 & \bb{65.00} & 10.00 & 61.67 & 1.70 & 48.33 & 10.00 & \br{76.67} & 30.00 & \br{66.67} & 10.00 & 66.67 & 3.30 & 60.00 & 16.67 & \bb{61.67} & 11.70 & 61.67 & 13.33 & 58.33 & 1.70 \\ \midrule
InternVL2.5-8B~\cite{internvl2024v25} & 39.65 & 59.09 & 40.76 & 9.80 & 38.67 & 59.74 & 40.13 & 10.15 & 36.82 & 60.49 & 39.35 & 12.58 & 39.12 & 59.40 & 40.49 & 9.84 & 37.63 & 60.28 & 39.56 & 11.70 & 38.31 & 60.20 & 39.68 & 10.05 \\
InternVL3-8B~\cite{internvl2025v3} & 40.78 & 57.94 & 41.85 & 0.85 & 39.80 & 58.85 & 40.86 & 0.46 & 36.89 & 61.42 & 38.37 & 0.41 & 40.19 & 58.49 & 41.22 & 0.75 & 37.88 & 60.63 & 39.14 & 0.50 & 39.44 & 59.24 & 40.49 & 1.02 \\
Llama3.2-11B~\cite{meta2024llama32vision11b} & 34.21 & 57.70 & 35.66 & 0.28 & 31.70 & 60.47 & 33.50 & 0.38 & 30.22 & 59.29 & 32.48 & 0.42 & 32.38 & 59.62 & 33.79 & 0.26 & 31.29 & 58.90 & 33.14 & 0.38 & 32.13 & 60.38 & 33.08 & 0.19 \\
Llama3-llava-next-8b~\cite{llava2024nextinterleave} & 30.89 & 50.61 & 36.81 & 0.42 & 30.40 & 52.62 & 36.51 & 0.44 & 29.81 & 51.08 & 37.30 & 1.63 & 30.49 & 51.12 & 36.45 & 0.58 & 29.88 & 51.46 & 36.67 & 0.90 & 30.55 & 51.16 & 36.47 & 0.65 \\
LLaVA-OneVision~\cite{llava2024onevision} & 33.90 & 63.18 & 35.19 & 0.52 & 32.66 & 63.42 & 34.15 & 0.87 & 31.04 & 60.76 & 34.40 & 2.33 & 32.95 & 62.97 & 34.76 & 1.05 & 31.66 & 62.53 & 34.05 & 1.16 & 32.74 & 62.95 & 34.55 & 1.01 \\
MiniCPM-O 2.6~\cite{openbmb2025minicpmo26} & 39.47 & 57.87 & 39.37 & 1.57 & 36.00 & 63.21 & 35.84 & 1.85 & 32.48 & 55.42 & 32.72 & 2.61 & 37.48 & 60.05 & 37.30 & 1.82 & 33.26 & 55.80 & 33.30 & 1.84 & 36.97 & 60.22 & 36.79 & 1.66 \\
MiniCPM-V 4.5~\cite{minicpmv2025_45} & 44.26 & 49.32 & 45.82 & 0.12 & 40.90 & 50.18 & 42.88 & 0.08 & 35.69 & 48.11 & 37.89 & 0.95 & 41.26 & 51.61 & 43.09 & 0.24 & 37.58 & 47.80 & 38.58 & 0.28 & 40.55 & 51.47 & 42.51 & 0.21 \\
Phi-4-multimodal~\cite{microsoft2025phi4mm} & 33.26 & 63.81 & 34.70 & 0.18 & 31.61 & 59.87 & 35.18 & 0.66 & 29.32 & 64.49 & 32.57 & 3.02 & 31.60 & 63.97 & 33.96 & 0.64 & 30.42 & 64.71 & 33.17 & 1.46 & 31.25 & 63.65 & 33.91 & 0.56 \\
Qwen3-VL-8B~\cite{qwen2025qwen3vl} & 50.56 & 46.21 & 52.61 & 0.63 & 45.43 & 50.52 & 49.00 & 0.87 & 44.46 & 46.50 & 52.18 & 3.87 & 47.70 & 46.94 & 51.80 & 1.30 & 46.91 & 47.89 & 51.03 & 1.73 & 46.64 & 47.93 & 51.03 & 1.05 \\ \midrule
Qwen2.5-VL-7B~\cite{qwen2025v25vl} & 49.99 & 21.36 & 60.66 & 0.17 & 40.61 & 15.72 & 62.38 & 0.61 & 42.16 & 16.46 & 63.26 & 1.74 & 45.37 & 20.22 & 60.73 & 0.26 & 44.85 & 20.08 & 61.21 & 0.46 & 45.00 & 19.01 & 61.21 & 0.23 \\
Qwen2.5-VL-7B-SFT & 58.43 & \br{1.73} & 43.71 & 0.08 & 50.72 & \br{4.14} & 50.12 & 0.12 & 49.79 & \br{1.11} & 50.83 & 0.06 & 53.21 & \br{1.97} & 48.45 & 0.05 & 52.89 & \br{1.56} & 48.53 & 0.06 & 51.31 & \br{1.66} & 49.79 & 0.08 \\
\midrule \midrule
\end{tabular}}
\end{table*}

\section{Experiments}\label{Evaluation}

\subsection{Experimental Setup}

\noindent \textbf{Training Details.}
We employ Qwen2.5-VL-7B~\cite{qwen2025v25vl} as the base model for all supervised fine-tuning experiments. Training parameters follow the default configuration of the official implementation~\cite{qwen2025v25vl}.
Depending on the question type, task-specific system prompts are dynamically loaded to guide the model. To preserve pretrained visual representations, the vision encoder and multimodal projector are frozen during fine-tuning. The model is trained for 10 epochs on 8$\times$ NVIDIA A100 GPUs.

\noindent \textbf{Compared Methods.}
All evaluations are conducted on our CTVid test set, covering Text-Centric Video Restoration and Text-Centric Multi-Quality VideoQA tasks. For restoration, we evaluate both classical quality-enhancement models and recent diffusion-based methods. For QA, we assess open-source and proprietary MLLMs.

\subsection{Evaluation on Text-Centric Video Restoration}
\noindent \textbf{Text-Centric Image Super-Resolution.}
We benchmark six state-of-the-art image SR methods on the CTVid test set, as shown in Table~\ref{tab:benchmark_SR}. The evaluated models include classical approaches~\cite{wang2021real,liang2021swinir} and recent diffusion-based techniques~\cite{wu2024seesr,wu2024one,chen2025adversarial,zhang2024degradation}. Our evaluation uses both perceptual image-quality metrics and text-aware measures. Strong perceptual scores do not guarantee textual fidelity. While SwinIR~\cite{liang2021swinir} and SeeSR~\cite{wu2024seesr} lead image-based methods according to perceptual metrics, S3Diff~\cite{zhang2024degradation} achieves the highest text detection and recognition accuracy.
Qualitative results in Fig.~\ref{fig:Vis_results} show that all models still introduce distortions in challenging text regions. Classical methods tend to over-smooth small or degraded characters, whereas diffusion-based models produce sharper structures but may hallucinate fine details. In the second example, SeeSR misrecognizes ``Fitness'' as ``Fliness'' and ``Wellness'' as ``Weiiness'', while SwinIR fails to produce legible letters. In the first example, the more complex Chinese characters undergo more severe degradation.

\noindent \textbf{Text-Centric Video Super-Resolution.}
We benchmark six state-of-the-art VSR models on the CTVid dataset, including classical methods~\cite{chan2022investigating,chan2022basicvsrpp,zhang2024realviformer} and recent diffusion-based approaches~\cite{yang2023mgldvsr,zhou2024upscale,chen2025dove}. 
As shown in Table~\ref{tab:benchmark_SR}, DOVE~\cite{chen2025dove} achieves the best video-based R$_{det}$ and F1$_{det}$ scores, reaching $63.36\%$ and $74.53\%$, respectively, and also obtains the highest video-based Acc$_{rec}$ of $44.72\%$. RealViFormer~\cite{zhang2024realviformer} obtains the best NED$_{rec}$ of 46.37\%, while BasicVSR++~\cite{chan2022basicvsrpp} obtains the highest P$_{det}$ of 91.11\%. Qualitative results in Fig.~\ref{fig:Vis_results} show that all models introduce some text distortion. Nevertheless, video-based methods can leverage temporal information to enhance consistency and legibility. In the first example, DOVE successfully restores complex Chinese characters with minimal visible distortion.

\noindent \textbf{Text-Centric Video Deblurring.}
We evaluate text-centric video deblurring using both image- and video-based state-of-the-art models~\cite{cho2021rethinking,chen2022simple,tsai2022stripformer,liang2022rvrt,Li_2023_CVPR}. The evaluation protocol mirrors that of the super-resolution tasks.
As shown in Table~\ref{tab:benchmark_deblur}, Stripformer~\cite{tsai2022stripformer} yields the best R$_{det}$, F1$_{det}$, and Acc$_{rec}$ scores of $69.99\%$, $78.43\%$, and $58.93\%$, respectively. ShiftNet~\cite{Li_2023_CVPR} yields the highest P$_{det}$ of 91.05\% and the lowest NED$_{rec}$ of 59.56\%, indicating stronger preservation of fine-grained structure under motion blur.
In the examples shown in Fig.~\ref{fig:Vis_results}, several methods recover text with only minor distortions. However, failures occur under large motion, including partial restoration and complete breakdown when motion exceeds model capacity.

\begin{figure}[htb]
    \centering
    \includegraphics[width=1.0\textwidth]{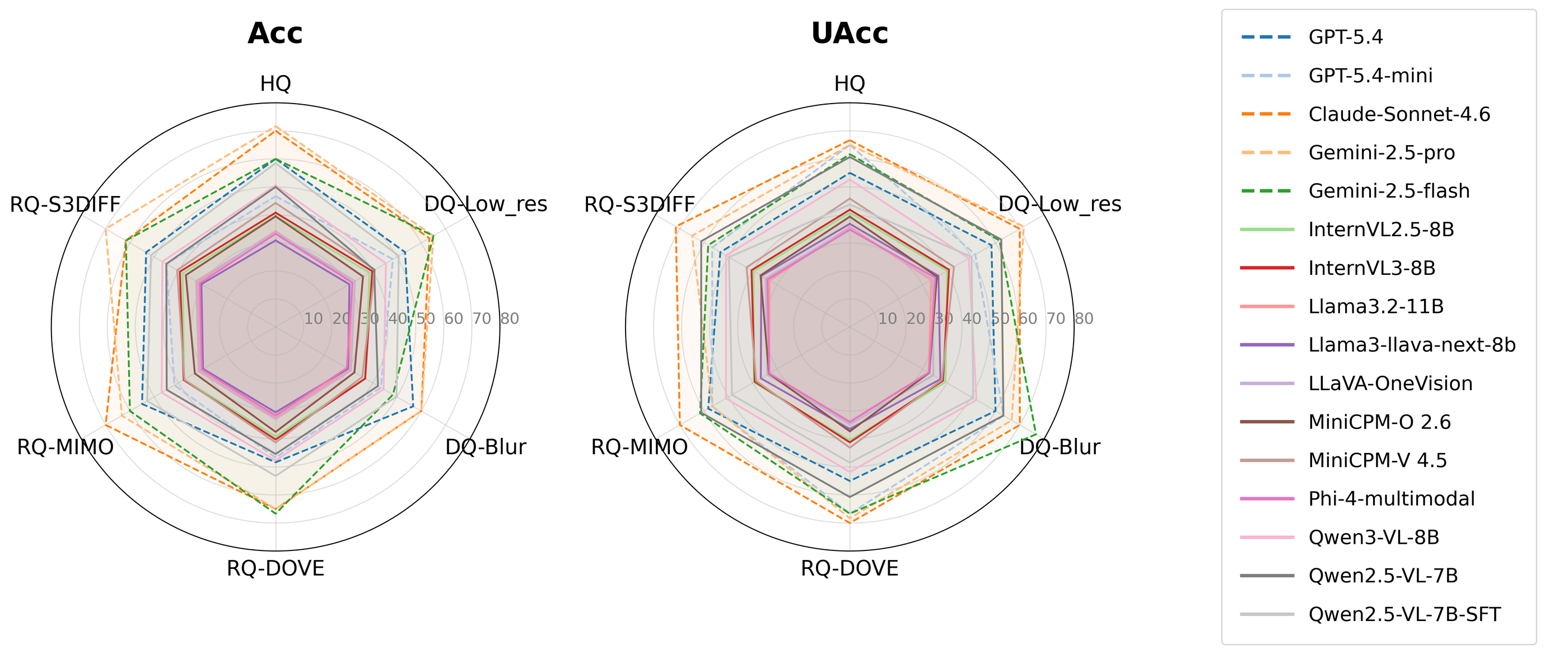}
    \caption{Accuracy (Acc) and uncertainty-aware accuracy (UAcc) radar plots for multiple MLLMs across six video-quality conditions. The top plot reports Acc, and the bottom plot reports UAcc.}
    \label{fig:radar}
\end{figure}

\subsection{Evaluation on Multi-Quality VideoQA}
\noindent \textbf{Text-Centric VideoQA for Spatial Understanding.}
We benchmark 16 MLLMs on the CTVid test set, including five proprietary models~\cite{openai2026gpt54,openai2026gpt54mini,anthropic2026claudesonnet,gemini2025v25} and ten open-source baselines~\cite{qwen2025v25vl,internvl2024v25,internvl2025v3,meta2024llama32vision11b,llava2024nextinterleave,llava2024onevision,openbmb2025minicpmo26,minicpmv2025_45,microsoft2025phi4mm,qwen2025qwen3vl}, together with our Qwen2.5-VL-7B-SFT variant.
As shown in Table~\ref{tab:all-imageQA_results}, proprietary MLLMs still achieve the strongest spatial VideoQA accuracy, but their robustness varies across quality conditions. Gemini-2.5-pro obtains the best accuracy on HQ, DQ-Low\_res, and RQ-S3DIFF inputs, with scores of 71.67\%, 65.00\%, and 70.00\%, respectively. Claude-Sonnet-4.6 leads on DQ-Blur and RQ-MIMO, reaching 60.00\% and 70.00\%, while Gemini-2.5-flash performs best on RQ-DOVE at 66.67\%. These results show that restoration does not yield a uniformly monotonic improvement: the best model can change with the restoration source, and perceptually enhanced videos may still alter text evidence in ways that affect downstream reasoning.
Among open-source models, Qwen3-VL-8B achieves the strongest zero-shot accuracy on HQ input (50.56\%), whereas Qwen2.5-VL-7B yields the highest open-source UAcc under all six conditions. After instruction tuning on ClearText-Video, Qwen2.5-VL-7B-SFT achieves the best open-source accuracy across every quality condition (58.43\%, 50.72\%, 49.79\%, 53.21\%, 52.89\%, and 51.31\%), improving over the Qwen2.5-VL-7B base model by 6.31--10.11 percentage points. The tuned model also substantially reduces overconfidence, but its lower UAcc indicates that accuracy gains and uncertainty-aware reliability do not always move together.
The radar chart in Fig.~\ref{fig:radar} summarizes these accuracy and UAcc trends across the six quality conditions. A controlled LR-vs-blur analysis further shows that blur is more damaging than low resolution: averaged over 16 MLLMs, LR reduces spatial QA accuracy by 3.14 points from HQ, whereas blur reduces it by 6.05 points. An OCR+LLM diagnostic baseline remains far below direct MLLM inference, suggesting that the task requires spatial grounding and multimodal reasoning beyond what is captured by extracted OCR text.

\noindent \textbf{Text-Centric VideoQA for Temporal Understanding.}
The temporal understanding VideoQA task follows the same input protocol as the spatial setting, using five input video-quality conditions that are uniformly 2$\times$ downsampled to satisfy memory constraints. To better assess temporal reasoning, we adopt a fixed downsampling scheme and minimize temporal subsampling whenever possible, thereby preserving temporal continuity across sequences. As shown in Table~\ref{tab:vqa_temporal_results}, Claude-Sonnet-4.6 consistently ranks first across all five temporal conditions. It obtains 60.02\% on HQ, 59.45\% on DQ-Blur, 59.34\% on RQ-MIMO, 58.88\% on RQ-S3DIFF, and 60.37\% on RQ-DOVE, and is the only model to exceed 58\% under any condition. In contrast to the spatial setting, the remaining proprietary models (Gemini-2.5-flash, Gemini-2.5-pro, and GPT-4o-mini) do not consistently outperform the open-source baselines in this setting: their scores fall into the high-40\% to low-50\% band, with Gemini-2.5-pro dropping to 45.83\% on HQ.
Among open-source models, performance is concentrated in the low-50\% range. Kimi-VL-16B achieves the best open-source accuracy on HQ, DQ-Blur, RQ-S3DIFF, and RQ-DOVE, while InternVL3-8B is marginally better on RQ-MIMO. Claude-Sonnet-4.6 consistently leads the strongest open-source model by about six percentage points, with the largest margin reaching 6.30 points on RQ-MIMO and the margin remaining above 5 points in all five temporal settings. This indicates that temporal text reasoning remains challenging even when individual frames contain readable text. Model rankings also vary across quality conditions: restoration improves some models but causes unstable changes for others, suggesting that temporal aggregation and restoration-induced artifacts both affect answer reliability.

\begin{table}[htb]
\centering
\caption{\textbf{Text-Centric Video Question Answering for Temporal Understanding.} We report accuracy across five video-quality conditions. The best and second-best results in each quality condition are highlighted in \textcolor{red}{\textbf{red}} and \textcolor{blue}{\textbf{blue}}, respectively, across all models.}
\label{tab:vqa_temporal_results}
\resizebox{1\linewidth}{!}{
\begin{tabular}{@{}cc|ccccc@{}}
\midrule
\midrule
\multicolumn{7}{c}{Accuracy\%($\uparrow$)}\\ \midrule
& Models & HQ & DQ-Blur~\cite{nah2019ntire} & RQ-MIMO~\cite{cho2021rethinking} & RQ-S3DIFF~\cite{zhang2024degradation} & RQ-DOVE~\cite{chen2025dove}
\\ \midrule
& Gemini-2.5-flash~\cite{gemini2025v25} & 51.55 & 51.20 & 52.12 & 51.32 & 53.04 \\
& Gemini-2.5-pro~\cite{gemini2025v25}   & 45.83 & 50.00 & 51.39 & 50.00 & 48.61 \\
& GPT-4o-mini~\cite{openai2024gpt4o}         & 50.06 & 50.52 & 50.29 & 50.13 & 49.94 \\
& Claude-Sonnet-4.6~\cite{anthropic2026claudesonnet} & \textcolor{red}{60.02} & \textcolor{red}{59.45} & \textcolor{red}{59.34} & \textcolor{red}{58.88} & \textcolor{red}{60.37} \\
\midrule
& Qwen2.5-VL-7B~\cite{qwen2025v25vl}    & 50.86 & 50.06 & 50.40 & 50.29 & 50.63 \\
& Kimi-VL-16B~\cite{team2025kimi}       & \textcolor{blue}{53.95} & \textcolor{blue}{53.49} & 52.92 & \textcolor{blue}{53.49} & \textcolor{blue}{54.18} \\
& InternVL3-8B~\cite{internvl2025v3}  & 52.92 & 52.46 & \textcolor{blue}{53.04} & 51.66 & 52.81 \\
& VideoLLaMA3-7B~\cite{zhang2025videollama} & 52.46 & 51.66 & 52.92 & 50.63 & 50.40 \\
\midrule
\end{tabular}%
}
\end{table}

\section{Discussion and Limitations}

\noindent \textbf{Restoration and understanding are coupled.}
CTVid does not treat restoration as an independent leaderboard detached from VideoQA. Instead, restoration constructs the RQ regime and enables a diagnostic question that prior single-quality TextVQA benchmarks cannot isolate: whether visually improving degraded text videos preserves the exact textual evidence required for downstream reasoning. Our results show that this relationship is not monotonic. Some restored videos become sharper but still alter character strokes, introduce plausible-looking text, or shift spatial cues, which can leave MLLMs less accurate than on degraded inputs. This supports evaluating restoration with both visual/text-fidelity metrics and downstream reasoning metrics.

\noindent \textbf{Controlled degradations and data release.}
Our DQ videos use controlled bicubic downsampling and motion blur to keep the underlying content, text annotations, and QA pairs fixed across HQ/DQ/RQ comparisons. This choice enables reproducible controlled analysis of resolution and blur, but it does not cover every artifact present in user videos, such as exposure changes, rolling shutter, sensor noise, ISP effects, and complex compression. Future versions of CTVid can extend the same paired protocol to richer real-world degradations. The release will include annotations, QA pairs, code, prompts, and permitted video assets and metadata so that these extensions can be reproduced under the same evaluation protocol.

\noindent \textbf{OCR probes and calibration.}
PaddleOCR is not used as ground truth. It initializes coarse text candidates during annotation and serves as a fixed OCR probe for reproducible text-fidelity measurement, while human annotators verify boxes, transcripts, QA pairs, and ambiguity. The OCR+LLM diagnostic baseline further confirms that extracted text alone is insufficient: CTVid still requires spatial grounding and multimodal reasoning. Finally, supervised fine-tuning on CTVid improves answer accuracy but does not uniformly improve reliability; lower UAcc and changed overconfidence indicate an accuracy-calibration trade-off that future models should address explicitly.

\section{Conclusion}

In this work, we introduced ClearText-Video (CTVid), which is, to our knowledge, the first large-scale, scene-text-aware video QA benchmark designed to examine how input video quality affects text-centric multimodal reasoning. CTVid is built around a controlled quality protocol: the same underlying text-rich videos are evaluated under High-Quality, Degraded-Quality, and Restored-Quality regimes. This design allows us to analyze how degradation and restoration affect model reasoning, which is difficult to study with existing single-quality TextVQA or restoration datasets.
Beyond dataset construction, CTVid defines a unified evaluation suite that connects low-level restoration with high-level video understanding. The benchmark supports text-centric image super-resolution, video super-resolution, video deblurring, and multi-quality VideoQA, together with human-verified text annotations, captions, and spatial/temporal QA pairs. This setup makes it possible to evaluate not only whether a method improves visual quality, but also whether it preserves the textual evidence needed for downstream reasoning.
Experiments with 18 restoration methods and 16 MLLMs show that current systems remain fragile under quality changes: restoration does not always improve reasoning accuracy, and visually enhanced videos can still alter textual evidence needed by MLLMs. We expect CTVid to support future research on text-faithful restoration, quality-robust multimodal reasoning, and evaluation protocols that jointly measure visual enhancement and text-grounded understanding.

\clearpage
\bibliographystyle{splncs04}
\bibliography{jinlong}

\clearpage
\appendix
\section{ClearText-Video Dataset}

\subsection{Diversity of Text-Centric Video Scenes}
\begin{figure*}[htb]
    \centering
    \includegraphics[width=\textwidth]{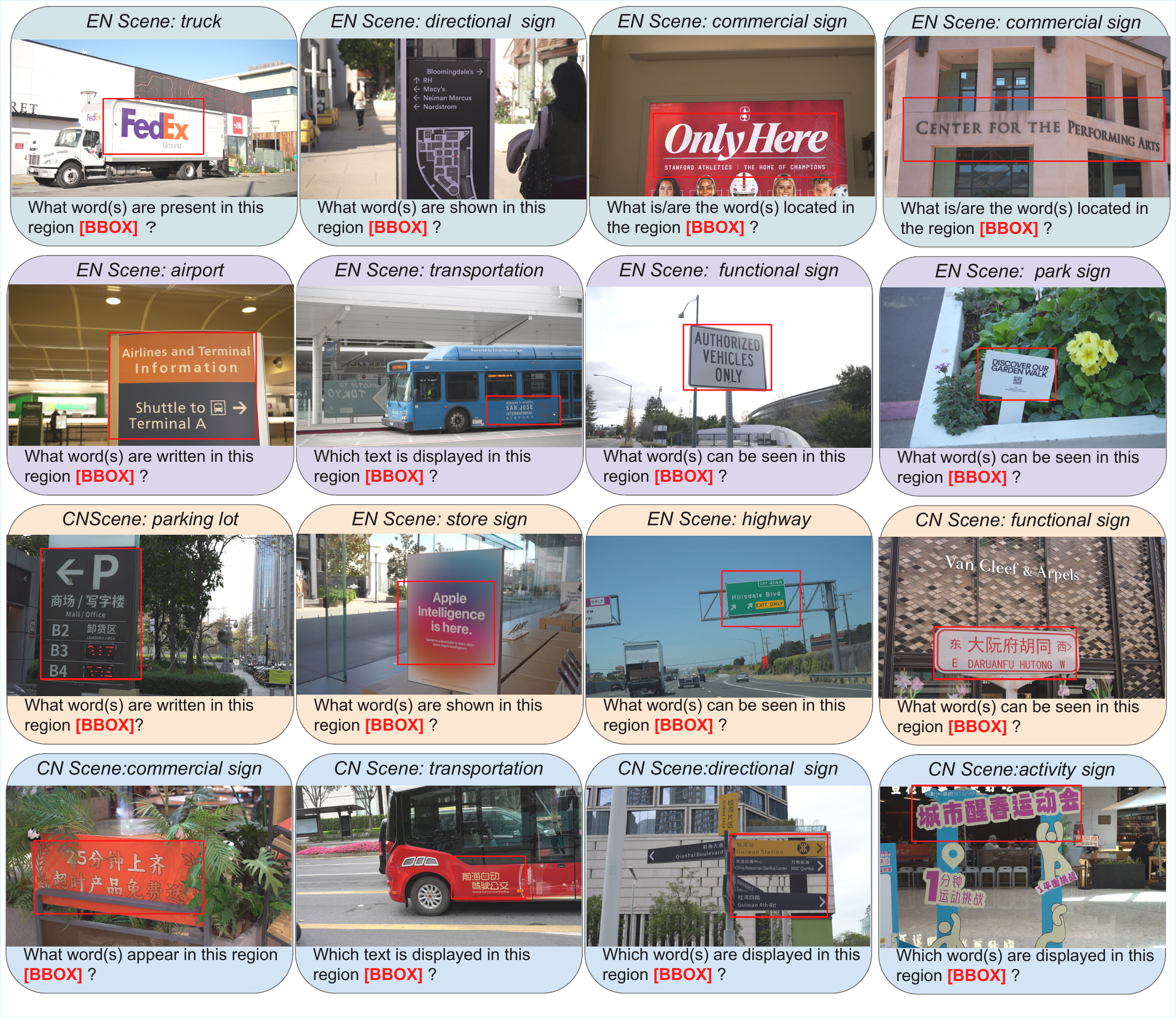}
    \caption{Example videos and their annotated questions from the ClearText-Video benchmark. Note: \br{[BBOX]} denotes bounding-box coordinates, visualized as the red detection boxes in the images. \textit{The examples span transportation, commercial, and public-service scenes in both English and Chinese environments, illustrating the multilingual and text-centric diversity of our benchmark. This diversity enables the evaluation of model generalization across varied real-world video scenarios.}}

    \label{fig:vis_videoQA_more}
\end{figure*}

\li{
ClearText-Video covers a wide range of text-centric video scenarios rather than a narrow set of canonical street-view scenes, as shown in Figure~\ref{fig:vis_videoQA_more}. The videos come from heterogeneous environments, including transportation hubs (\textit{e.g.}, airports, bus terminals, and highways), urban driving scenes (\textit{e.g.}, trucks, buses, and roadside signs), commercial areas (\textit{e.g.}, storefronts, billboards, and activity banners), and public facilities (\textit{e.g.}, functional and directional signs in parks and parking lots). In these settings, textual content is a primary carrier of semantic information and is closely linked to navigation, transportation, and commercial guidance.
}

\li{
To further challenge models, the benchmark includes both English and Chinese scenes with diverse layouts and sign types, ranging from large outdoor billboards and vehicle liveries to small, densely packed information boards. Text appears under varying viewpoints, distances, and levels of background clutter, and individual text instances are localized with fine-grained bounding boxes over time. This combination of multilingual, multi-domain, and structurally diverse text instances supports evaluation across languages, scene categories, and visual conditions rather than emphasizing the biases of a particular dataset.
}

\li{
Overall, the diversity of ClearText-Video makes it suitable not only for evaluating text spotting and reading in videos, but also for studying higher-level video-language understanding in realistic, text-rich environments. By exposing models to a broad spectrum of real-world uses of text in videos, our benchmark provides a more faithful approximation of practical deployment scenarios.
}

\begin{figure}[htb]
    \centering
    \includegraphics[width=0.55\linewidth]{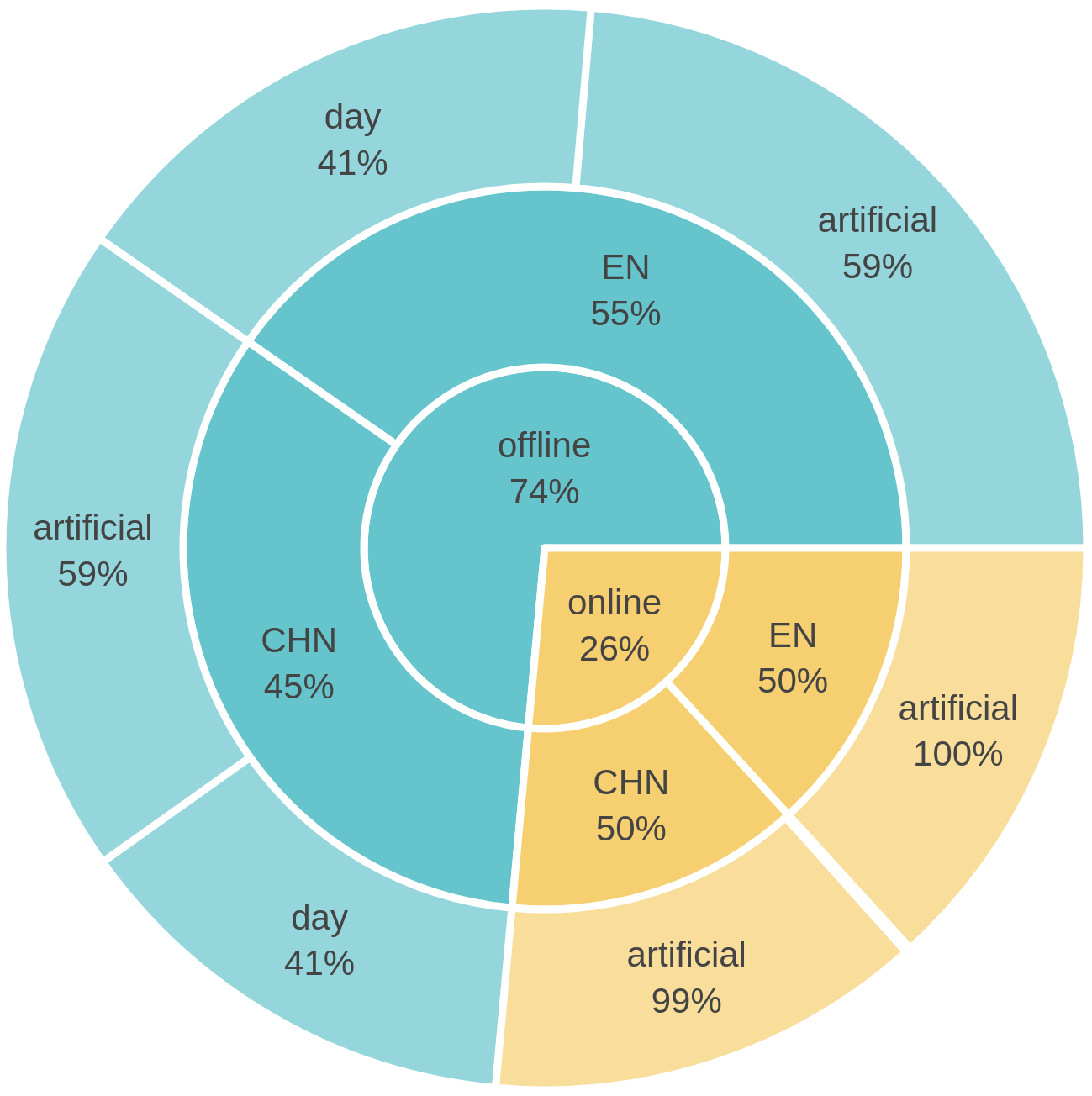}
    \caption{\li{Sunburst visualization of the ClearText-Video test set composition. 
    From the center outward, the rings represent video source (offline vs.\ online), language (EN vs.\ CHN), and lighting condition (artificial vs.\ daylight). 
Offline videos constitute 74\% of the test set, with a balanced distribution of English and Chinese clips and a mix of artificial and natural lighting. The online portion (26\%) is evenly split between English and Chinese and is almost entirely captured under artificial lighting.}}

    \label{fig:test_set_origins}
\end{figure}

\subsection{Test Set Details}
\li{
When constructing our test set, we prioritize high-quality (HQ) videos that are well illuminated, clearly show text regions, and avoid excessive global motion blur. The resulting 312 HQ videos are split equally between English and Chinese scenes. Three-quarters of the test set comes from offline recordings, while the remaining quarter is collected from online sources. Among offline videos, roughly 60\% of the clips are captured under artificial lighting (\textit{e.g.}, indoor scenes or outdoor scenes at night), and the remaining 40\% under natural lighting, as shown in Figure~\ref{fig:test_set_origins}. For online videos, almost all clips are artificially lit, reflecting the predominance of indoor and nighttime footage in user-generated content.
}
\li{
To systematically study robustness to quality degradation, we further derive video-quality variants from each HQ test video. Following prior work~\cite{nah2019ntire}, we apply bicubic downscaling ($\times 4$) to obtain low-resolution (LR) versions and local motion blur to obtain blurry versions. This yields matched sets of LR and blurry videos with the same number of samples as the HQ test set, enabling controlled evaluations across three quality conditions (HQ, LR, and blur) under identical content and question distributions.
}

\subsection{Release, Licensing, and Dataset Comparison}

CTVid will be released with train/test splits, human-verified text boxes and transcripts, captions, temporal trajectories, spatial/temporal QA pairs, degradation-generation scripts, evaluation code, prompts, and baseline outputs. Self-captured clips and their annotations will be distributed under a research license. For online clips with redistribution restrictions, we will release source IDs or URLs where permitted, clip ranges, annotations, and preprocessing scripts. Raw videos will be redistributed only when compatible licenses or explicit permissions allow it.

Table~\ref{tab:dataset_comparison_expanded} complements the comparison in the main text by focusing on text-centric VQA and video-OCR benchmarks. CTVid differs from prior datasets by jointly providing large-scale QA, dense frame-level text boxes and transcripts, bilingual scenes, paired HQ/DQ/RQ inputs, and restoration-aware evaluation. Here, paired HQ/DQ/RQ denotes content-matched high-quality, degraded-quality, and restored-quality videos. Restoration-aware evaluation measures downstream reasoning performance under restoration-induced quality variation.

\begin{table}[t]
\caption{Comparison with text-centric video benchmarks. ``\#Boxes / Trans.'' denotes text bounding boxes, OCR tokens, or transcripts when explicitly provided. ``Dense Frame-Text'' indicates dense frame-level text annotations rather than sparse or question-grounded evidence. ``Partial'' indicates related but incomplete support.}
\centering
\label{tab:dataset_comparison_expanded}
\resizebox{0.99\textwidth}{!}{%
\begin{tabular}{@{}ccccccccc@{}}
\toprule
\textbf{Dataset}
& \textbf{\#QA}
& \makecell{\textbf{\#Boxes /}\\\textbf{Trans.}}
& \makecell{\textbf{Spatial}\\\textbf{QA}}
& \makecell{\textbf{Temporal}\\\textbf{QA}}
& \makecell{\textbf{Bilingual/}\\\textbf{Multilingual}}
& \makecell{\textbf{Dense}\\\textbf{Frame-Text}}
& \makecell{\textbf{Paired}\\\textbf{HQ/DQ/RQ}}
& \makecell{\textbf{Rest.}\\\textbf{Eval.}} \\
\midrule
TextVQA~\cite{singh2019towards}
& 45,336
& \makecell{OCR tokens\\\#boxes: NA}
& \xmark
& \xmark
& \xmark
& \xmark
& \xmark
& \xmark \\
NewsVideoQA~\cite{jahagirdar2023watching}
& 8,672
& \makecell{OCR tokens\\+ subtitles}
& \xmark
& \cmark
& \xmark
& \xmark
& \xmark
& \xmark \\
EgoTextVQA~\cite{zhou2025egotextvqa}
& 7,064
& \makecell{OCR stats only\\\#boxes: NA}
& Partial
& \cmark
& \xmark
& \xmark
& \xmark
& \xmark \\
ViTXT-GQA~\cite{zhou2024scene}
& 2,055
& \makecell{52,494 boxes\\2,227 segments}
& \cmark
& \cmark
& \xmark
& Partial
& \xmark
& \xmark \\
M4-ViteVQA~\cite{zhao2022towards}
& 25,123
& \makecell{OCR tokens\\avg. 56.92/video}
& Partial
& \cmark
& \xmark
& \xmark
& \xmark
& \xmark \\
MME-VideoOCR~\cite{shi2025mme}
& 2,000
& NA
& \cmark
& \cmark
& Partial
& \xmark
& \xmark
& \xmark \\
\midrule
\textbf{ClearText-Video (Ours)}
& \textbf{220K+}
& \makecell{\textbf{1.6M boxes}\\\textbf{+ transcripts}}
& \cmark
& \cmark
& \cmark
& \cmark
& \cmark
& \cmark \\
\bottomrule
\end{tabular}
}
\end{table}

\subsection{Question Answering Generation Details}

\subsubsection{Text-Centric VideoQA for Spatial Understanding}
\jiaming{
This task targets scene-text comprehension in a spatial context. For each frame, we generate three types of spatially grounded questions: (1) multiple-choice questions that ask which text appears within a specified bounding-box region, (2) true/false questions that verify whether a specific text label exists within the region, and (3) fill-in-the-blank questions that require identifying all text labels within the bounding box.
Our generation pipeline combines rule-based spatial reasoning with LLM-based question formulation. First, an automated spatial partitioning algorithm analyzes the text distribution within each frame. It selects either vertical or horizontal partition lines to divide the image into regions where text labels are spatially separated. This process determines both the target bounding box (defined as the minimum enclosing rectangle of selected text polygons) and the corresponding ground-truth labels. To ensure question diversity and prevent spatial overlap, we maintain a question-history buffer that guides the LLM to generate varied question phrasings while incorporating the exact bounding-box coordinates. Each generated question undergoes validation through an \textit{AnswerChecker} module, which checks whether (a) the question correctly references the bounding box, (b) the spatial constraints are logically consistent, and (c) the expected answer format is properly structured.
To create challenging distractors for multiple-choice questions, we employ a hybrid similarity-based option generation system. The \textit{UnifiedCandidateGenerator} produces visually and semantically confusing alternatives through three mechanisms: OCR-level confusion (character-shape similarity), semantic confusion (contextually related terms), and visual confusion (spatially adjacent text). This design requires careful spatial reasoning rather than simple text matching. The final option sets are shuffled to randomize answer positions.
}

\begin{figure*}[htb]
    \centering
    \includegraphics[width=0.85\linewidth]{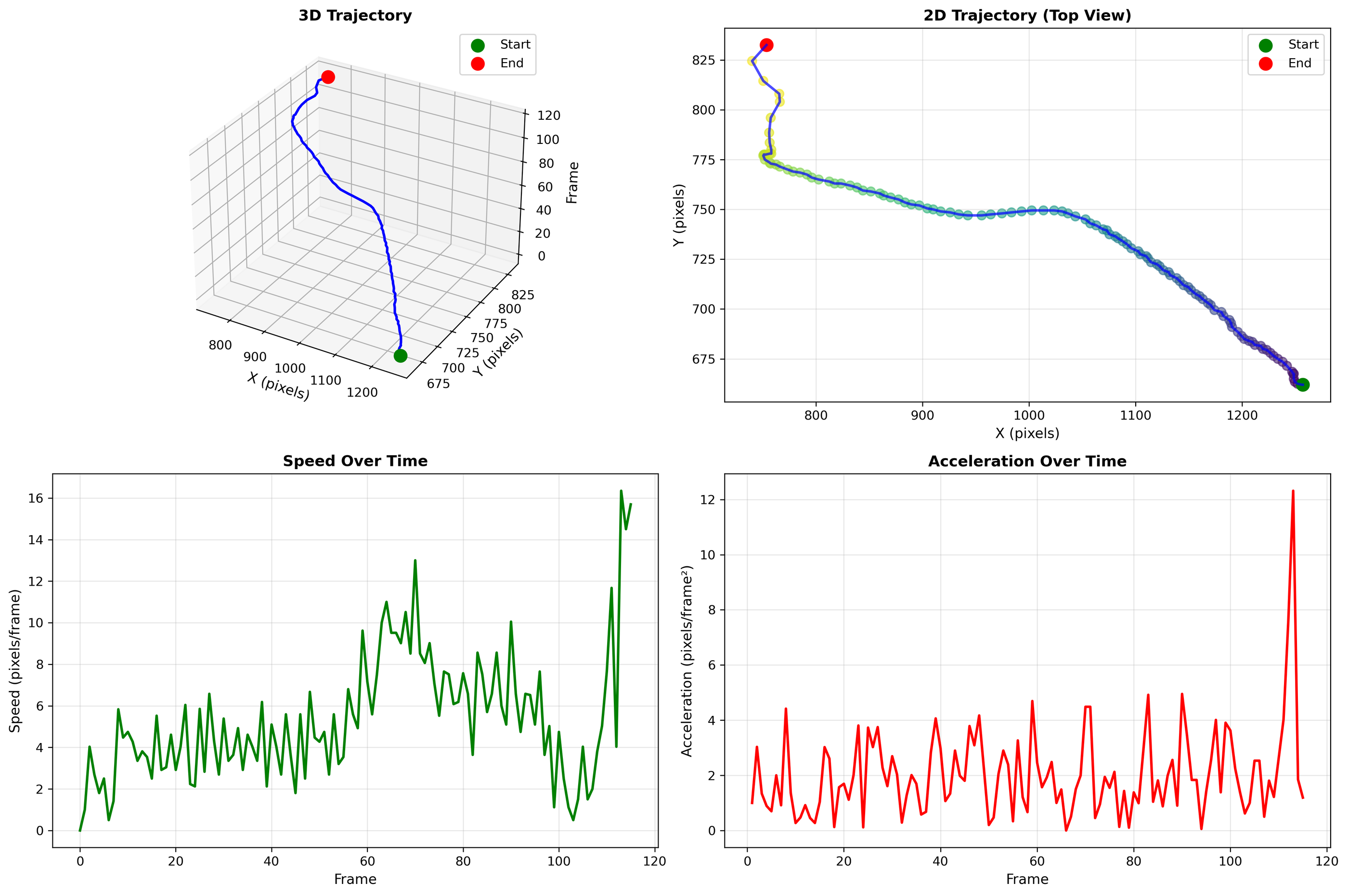}
    \caption{An example text instance showing its trajectory, speed, and acceleration over time.}
    \label{fig:trajectory}
\end{figure*}

\begin{figure*}[htb]
    \centering
    \includegraphics[width=0.85\linewidth]{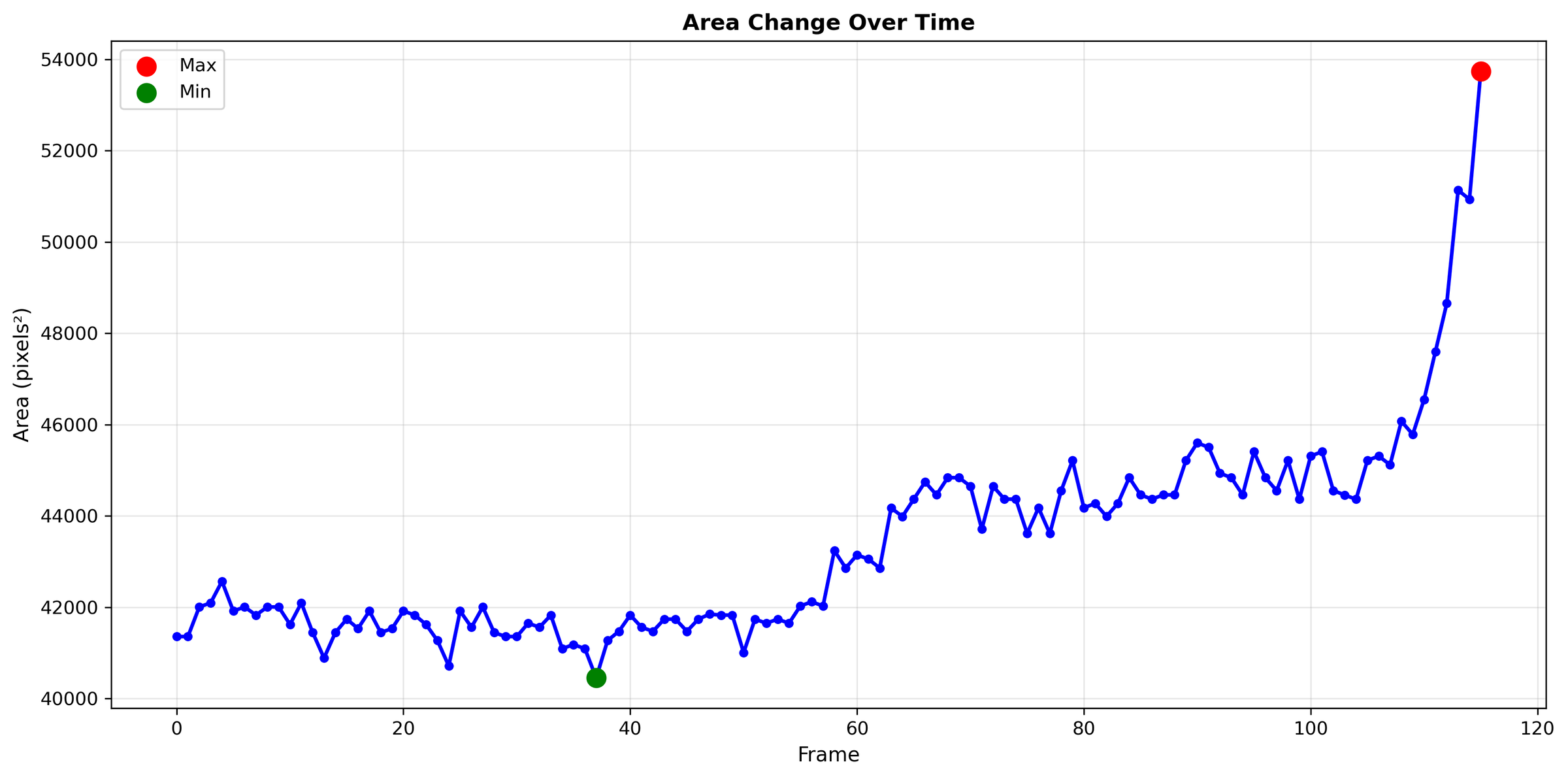}
    \caption{Temporal variation of the bounding-box area for the same text instance.}
    \label{fig:area_change}
\end{figure*}

\subsubsection{Text-Centric VideoQA for Temporal Understanding}
\jiaming{
This task evaluates temporal reasoning about scene-text motion and visibility patterns across frames in video clips. The questions are generated by analyzing the trajectories of text polygons. For each text instance, we extract frame-wise center coordinates, areas, speeds (Euclidean displacement between consecutive frames), and accelerations. Figure~\ref{fig:trajectory} shows an example text instance, including its trajectory across frames together with the corresponding speed and acceleration curves. Figure~\ref{fig:area_change} illustrates the temporal variation of the bounding-box area, highlighting how the apparent scale of the text instance evolves over time.
These temporal features feed into five specialized question generators: (1) Presence \& Visibility, which queries whether text appears consistently, how many frames it is visible in, disappearance counts, first/last appearance frames, and overall appearance patterns (\textit{e.g.}, ``always present''); (2) Spatial Localization, which identifies text regions at first/last appearance, the most frequent region, frames in which the text crosses the horizontal center line, and region traversal counts; (3) Motion \& Trajectory, which determines the main motion direction and frames with maximum speed/acceleration; (4) Size \& Scale, which determines text size and its temporal variation trends (increasing or decreasing); and (5) Boundary Interaction, which detects whether text touches frame edges during motion.
All questions are generated using deterministic rules based on statistical thresholds. The generators produce structured QA items comprising multiple-choice options with plausible distractors, numerical answers, and categorical labels for directions and patterns. Human annotators then screen and refine the automatically generated questions so that only meaningful, high-quality items are retained. This hybrid pipeline combines the consistency and reproducibility of rule-based generation with human verification, reducing errors in automatically generated items.
}

\section{Additional Experimental Details}
\li{This section provides additional details about the experimental setup for the ClearText-Video benchmark. 
We first summarize the image and video super-resolution and deblurring models used as low-level restoration baselines. We then describe the multimodal large language models (MLLMs) evaluated in the spatial and temporal VideoQA experiments and qualitative analyses.
}

\subsection{Details of Image Super-Resolution Models}

\noindent \textbf{Real-ESRGAN~\cite{wang2021real}}:
Real-ESRGAN extends the ESRGAN framework to practical real-world blind super-resolution by training solely on synthetically degraded data generated via a high-order degradation model that more faithfully simulates complex real degradations such as noise, blur, compression, and ringing.
It explicitly models common ringing and overshoot artifacts using sinc filters and employs a U-Net discriminator with spectral normalization to provide strong per-pixel adversarial supervision and stabilize training.
Experiments on multiple real-world datasets report improved visual quality over previous blind SR methods while retaining an efficient on-the-fly synthetic data-generation pipeline.

\noindent \textbf{SwinIR~\cite{liang2021swinir}}:
SwinIR is a strong image restoration baseline that replaces convolutional backbones with Swin Transformer blocks, leveraging shifted-window self-attention to jointly capture local and non-local dependencies.
The network consists of shallow feature extraction, deep feature extraction via residual Swin Transformer blocks, and a reconstruction module, and is instantiated for super-resolution, image denoising, and JPEG compression artifact removal.
The reported results show that SwinIR outperforms prior CNN-based methods on these tasks while reducing the parameter count by up to 67\%.

\noindent \textbf{SeeSR~\cite{wu2024seesr}}:
SeeSR tackles real-world image super-resolution with a semantic-aware framework that exploits pretrained text-to-image diffusion priors to better preserve semantic fidelity under severe degradations.
A degradation-aware prompt extractor is trained to generate complementary hard (tag-like) and soft semantic prompts from low-resolution inputs, guiding the diffusion model to synthesize detailed and semantically consistent high-resolution images.
During inference, the low-resolution image is further injected into the initial sampling noise to suppress spurious random details, which the authors report improves texture realism and semantic preservation.

\noindent \textbf{OSEDiff~\cite{wu2024one}}:
OSEDiff is a one-step diffusion network for real-world image super-resolution that utilizes a pretrained text-to-image diffusion model as both a generator and a regularizer.
Instead of starting diffusion from random noise, OSEDiff takes the low-quality image as the initial state and fine-tunes a subset of trainable layers to handle complex real degradations, thereby reducing stochastic uncertainty in the reconstruction.
Variational score distillation in latent space aligns the one-step model with multi-step diffusion priors. The authors report that this design allows OSEDiff to achieve competitive Real-ISR quality with a single sampling step.

\noindent \textbf{S3Diff~\cite{zhang2024degradation}}:
S3Diff proposes a degradation-guided one-step image super-resolution model that enhances a pretrained diffusion prior with explicit degradation modeling.
It introduces a degradation-guided Low-Rank Adaptation (LoRA) module, which adjusts model parameters conditioned on degradation estimates from a pretrained degradation network, while preserving the generative prior of the underlying diffusion model.
An online negative-sample generation strategy and classifier-free guidance are further incorporated during training and inference to improve perceptual realism, yielding a data- and degradation-dependent SR model with high efficiency.

\noindent \textbf{AdcSR~\cite{chen2025adversarial}}:
AdcSR is a Real-ISR method developed under the Adversarial Diffusion Compression framework, which distills the one-step diffusion model OSEDiff into a structurally compressed diffusion-GAN.
The method removes components such as the VAE encoder and text- and time-conditioning modules. It then prunes feature channels in the U-Net and VAE decoder while preserving network depth and performs feature-level knowledge distillation with an adversarial loss.
This design yields a compact PixelUnshuffle--U-Net--decoder architecture that reduces parameters, computation, and inference latency while maintaining competitive image quality relative to previous SD-based one-step Real-ISR approaches.

\subsection{Details of Video Super-Resolution Models}
\noindent \textbf{RealBasicVSR~\cite{chan2022investigating}}:
RealBasicVSR is a real-world video super-resolution (VSR) method that explicitly studies the trade-offs between exploiting long-term temporal propagation and avoiding artifact amplification under complex in-the-wild degradations. 
It augments BasicVSR with a lightweight image cleaning module that preprocesses each frame to remove degradations before propagation, together with a dynamic refinement scheme that iteratively applies the cleaning module at test time based on a stopping criterion. 
Reported experiments on VideoLQ and other real-world benchmarks show improved perceptual quality and efficiency over prior blind and data-augmentation-based VSR approaches.

\noindent \textbf{BasicVSR++~\cite{chan2022basicvsrpp}}:
BasicVSR++ improves upon BasicVSR by introducing second-order grid propagation and flow-guided deformable alignment to more effectively aggregate spatiotemporal information. 
The proposed grid propagation relaxes the first-order Markov assumption and performs aggressive bidirectional, second-order propagation to repeatedly refine features across frames, enhancing robustness in occluded and fine-detail regions. 
Flow-guided deformable alignment uses optical flow as the base offsets for deformable convolutions, stabilizing training while retaining offset diversity. 
With these changes, BasicVSR++ reports gains over BasicVSR on multiple VSR benchmarks, including REDS4, with a similar parameter budget.

\noindent \textbf{RealViFormer~\cite{zhang2024realviformer}}:
RealViFormer is a recurrent Transformer network for real-world VSR that compares spatial and channel attention as temporal aggregation mechanisms under real degradations. 
It adopts channel attention as the primary temporal fusion strategy via a Channel Attention Fusion (CAF) module and introduces an Improved Channel Attention (ICA) block that combines squeeze--excite and covariance-based channel rescaling for better high-frequency reconstruction. 
With these components, RealViFormer reports strong performance on real-world datasets (VideoLQ and RealVSR) and synthetic benchmarks while using fewer parameters and achieving lower runtime than RealBasicVSR.

\noindent \textbf{MGLD-VSR~\cite{yang2023mgldvsr}}:
MGLD-VSR is a real-world VSR algorithm that leverages pretrained latent diffusion models and explicitly addresses temporal consistency through motion guidance. 
It exploits temporal dynamics in low-resolution videos to guide the diffusion sampling path with a motion-guided loss that encourages coherent content across generated high-resolution frames.
To further suppress temporal discontinuities, MGLD-VSR adds temporal modules to the decoder and optimizes them with a sequence-oriented loss. The paper reports improved perceptual quality on real-world VSR benchmarks.

\noindent \textbf{Upscale-A-Video~\cite{zhou2024upscale}}:
Upscale-A-Video is a text-guided latent diffusion framework for real-world video super-resolution built on a pretrained Stable Diffusion $\times4$ upscaler. 
To mitigate temporal instability from stochastic diffusion sampling, it uses a local--global strategy that combines temporal U-Net layers on short video segments with recurrent latent propagation for long-range refinement. 
Conditioned on low-resolution video and optional text prompts, the model is designed to improve temporal consistency in upscaled videos; the authors report competitive performance on real-world VSR benchmarks.

\noindent \textbf{DOVE~\cite{chen2025dove}}:
DOVE is an efficient one-step diffusion model for real-world video super-resolution obtained by fine-tuning a pretrained video diffusion model (CogVideoX) for the VSR task. 
To make single-step VSR training feasible, it introduces a latent-pixel training strategy with a two-stage scheme that gradually adapts the diffusion model from latent-domain supervision to pixel-domain fidelity. 
The authors also construct the HQ-VSR dataset and report that DOVE achieves competitive restoration quality relative to multi-step diffusion-based VSR methods, with up to a $28\times$ speedup over methods such as MGLD-VSR.

\subsection{Details of Deblurring Models}
\noindent \textbf{MIMO-UNet+~\cite{cho2021rethinking}}:
MIMO-UNet+ is an enhanced multi-input multi-output U-Net architecture that revisits the coarse-to-fine paradigm for single-image deblurring using a single encoder--decoder network~\cite{cho2021rethinking}. 
It takes blurry images at multiple resolutions as inputs and produces deblurred results at corresponding scales, while an asymmetric feature fusion module aggregates cross-scale features.
The model is trained on the GoPro dynamic scene deblurring dataset~\cite{nah2017deep} and the RealBlur real-world blur dataset~\cite{rim2020real}.

\noindent \textbf{NAFNet~\cite{chen2022simple}}:
NAFNet (Nonlinear Activation Free Network) is a simple image restoration baseline that removes conventional nonlinear activations and instead relies on linear and multiplicative operations while preserving strong representational capacity~\cite{chen2022simple}. 
Its architecture is built from lightweight residual blocks with channel attention, making it efficient and scalable to high-resolution inputs.
For motion deblurring, NAFNet is commonly trained on the GoPro dataset~\cite{nah2017deep}.

\noindent \textbf{Restormer~\cite{zamir2022restormer}}:
Restormer is a transformer-based architecture for high-resolution image restoration that introduces multi-Dconv head transposed attention (MDTA) and a gated-Dconv feed-forward network to model both local and long-range dependencies~\cite{zamir2022restormer}. 
Its encoder--decoder structure operates directly on full-resolution feature maps without window partitioning, making it suitable for large images in restoration tasks. 
For single-image motion deblurring, Restormer is typically trained on GoPro~\cite{nah2017deep} and applied to benchmarks such as HIDE and RealBlur-J/RealBlur-R~\cite{rim2020real}.

\noindent \textbf{Stripformer(GoPro)~\cite{tsai2022stripformer}}:
Stripformer is a transformer-based deblurring model that employs stripe-wise self-attention to efficiently capture long-range spatial interactions with linear complexity in image resolution~\cite{tsai2022stripformer}. 
The GoPro configuration focuses on dynamic scene deblurring, where motion blur is synthesized from high-speed videos.
In this setting, Stripformer is trained on the GoPro dataset~\cite{nah2017deep}.

\noindent \textbf{Stripformer(RealBlur-J)~\cite{tsai2022stripformer}}:
Stripformer(RealBlur-J) uses the same stripe-based transformer architecture but adapts it to real-world blur in the JPEG domain.
It is trained or fine-tuned on the RealBlur-J subset of the RealBlur dataset, which provides paired blurred and sharp images processed through the camera ISP pipeline~\cite{rim2020real}.
This configuration targets realistic deblurring scenarios where images are stored and processed as compressed JPEGs.

\noindent \textbf{Stripformer(RealBlur-R)~\cite{tsai2022stripformer}}:
The RealBlur-R configuration of Stripformer applies the same stripe-wise transformer design to images in the camera raw domain.
It is trained on the RealBlur-R subset of the RealBlur dataset, which provides blurred and sharp image pairs in RAW format~\cite{rim2020real}. 
This setup enables the model to operate directly on raw sensor data before in-camera processing.

\noindent \textbf{RVRT(DVD)~\cite{liang2022rvrt}}:
RVRT (Recurrent Video Restoration Transformer) is a unified recurrent transformer framework that processes video clips sequentially and reuses features over time for video restoration~\cite{liang2022rvrt}. 
It combines local windowed self-attention with recurrent hidden states to capture both short- and long-range temporal dependencies in degraded sequences.
In the DVD configuration, RVRT is trained on the Deep Video Deblurring (DVD) dataset from Su et al.~\cite{su2017deep}, which is constructed from high-frame-rate videos with synthetically generated motion blur.

\noindent \textbf{RVRT(GoPro)~\cite{liang2022rvrt}}:
RVRT(GoPro) uses the RVRT framework for video deblurring on sequences derived from the GoPro dynamic scene dataset~\cite{nah2017deep}. 
Short video clips are formed by grouping consecutive frames, and the recurrent transformer aggregates temporal information across these frames.
This configuration focuses on leveraging the GoPro dataset for learning spatially varying motion blur patterns in video.

\noindent \textbf{ShiftNet(DVD)~\cite{Li_2023_CVPR}}:
ShiftNet is a video restoration framework built on grouped spatiotemporal shift blocks, which shift feature channels across time and space and then fuse them through standard 2D convolutions~\cite{Li_2023_CVPR}. 
These shift operations provide a large effective receptive field for multi-frame aggregation without explicit optical flow or heavy transformer modules.
In the DVD setting, ShiftNet is trained on the Deep Video Deblurring dataset~\cite{su2017deep} for real video deblurring.

\noindent \textbf{ShiftNet(GoPro)~\cite{Li_2023_CVPR}}:
ShiftNet(GoPro) applies the same grouped spatiotemporal shift architecture to video sequences constructed from the GoPro dataset~\cite{nah2017deep}. 
By shifting and fusing feature maps across neighboring frames, it implicitly models inter-frame motion for dynamic scene deblurring.
This configuration uses GoPro-based sequences to learn typical hand-held camera motion blur in videos.

\subsection{Details of MLLMs}

The final quantitative evaluations use task-specific model sets. The spatial VideoQA evaluation in Table~\ref{tab:all-imageQA_results} includes 16 MLLMs: five proprietary models, ten open-source baselines, and our Qwen2.5-VL-7B-SFT variant. The temporal VideoQA evaluation in Table~\ref{tab:vqa_temporal_results} reports an eight-model subset. Grok-4-fast is not included in the quantitative tables and is retained only for temporal qualitative visualizations.

\noindent \textbf{GPT-5.4~\cite{openai2026gpt54}}:
GPT-5.4 is the larger of the two OpenAI models included in our spatial quantitative evaluation in Table~\ref{tab:all-imageQA_results}.
We use it as a proprietary general-purpose MLLM baseline for text-centric spatial VideoQA, where the task requires accurate text reading, spatial grounding, and calibrated answering across multiple video-quality conditions.

\noindent \textbf{GPT-5.4-mini~\cite{openai2026gpt54mini}}:
GPT-5.4-mini is the compact member of the same model family included in the spatial quantitative evaluation.
It provides a smaller, higher-throughput proprietary baseline for evaluating whether compact models preserve robustness under degradation and restoration.

\noindent \textbf{GPT-4o-mini~\cite{openai2024gpt4o}}:
GPT-4o-mini is a smaller member of the GPT-4o family that supports textual and multimodal reasoning.
In the final tables, it is used for temporal VideoQA rather than the spatial evaluation, providing a compact proprietary baseline for comparisons with larger Gemini and Claude models.

\noindent \textbf{Grok-4-fast~\cite{xai2025grok4fast}}:
Grok-4-fast is an xAI reasoning model optimized for lower-cost inference.
It is not included in the final quantitative tables; in this appendix, Grok-4-fast is retained only in temporal qualitative visualizations to illustrate model-specific failure modes under multi-quality inputs.

\noindent \textbf{Claude-Sonnet-4.6~\cite{anthropic2026claudesonnet}}:
Claude-Sonnet-4.6 is the Anthropic proprietary MLLM included in both final spatial and temporal VideoQA tables.
It serves as a strong reasoning-oriented baseline and is particularly useful for comparing accuracy with uncertainty behavior across quality-altered videos.

\noindent \textbf{Gemini-2.5-pro~\cite{gemini2025v25}}:
Gemini-2.5-pro is a Google DeepMind model designed for complex multi-step reasoning and multimodal analysis.
We evaluate it in both VideoQA settings as the larger model in the Gemini 2.5 family.

\noindent \textbf{Gemini-2.5-flash~\cite{gemini2025v25}}:
Gemini-2.5-flash is a lightweight Gemini model optimized for low-latency inference.
We include it in both VideoQA settings to evaluate the accuracy and robustness of a higher-throughput proprietary model.

\noindent \textbf{InternVL2.5-8B~\cite{internvl2024v25}}:
InternVL2.5-8B is an 8-billion-parameter multimodal large language model from the InternVL 2.5 family that builds upon the InternVL 2.0 architecture.
It introduces refined training and evaluation strategies, together with higher-quality data curation, to improve both perception and reasoning over images and text.
In our benchmark, it is used as an open-source baseline for the spatial VideoQA table.

\noindent \textbf{InternVL3-8B~\cite{internvl2025v3}}:
InternVL3-8B is an 8-billion-parameter model from the InternVL3 family that succeeds InternVL2.5.
It extends the capabilities of the series to application scenarios such as tool-augmented agents, GUI understanding, industrial inspection, and 3D vision perception.
The model is included in both the spatial and temporal quantitative evaluations as a representative open-source MLLM.

\noindent \textbf{Llama3.2-11B~\cite{meta2024llama32vision11b}}:
Llama3.2-11B denotes the 11-billion-parameter member of Meta's Llama 3.2 Vision family of multimodal models.
It is an instruction-tuned image reasoning model that accepts both text and image inputs and generates text outputs.
We include it as an open-source spatial VideoQA baseline in Table~\ref{tab:all-imageQA_results}.

\noindent \textbf{Llama3-llava-next-8b~\cite{llava2024nextinterleave}}:
Llama3-llava-next-8b is an 8-billion-parameter LLaVA-NeXT model obtained by fine-tuning Meta-Llama-3-8B-Instruct with multimodal instruction-following data.
It combines a transformer language backbone with a vision encoder and a learned projector, following the LLaVA design for end-to-end image--text understanding.
We evaluate it as an open-source spatial VideoQA baseline to cover the LLaVA-NeXT model family.

\noindent \textbf{LLaVA-OneVision~\cite{llava2024onevision}}:
LLaVA-OneVision is a family of large multimodal models that unifies single-image, multi-image, and video understanding within a single architecture.
The design consolidates insights from the LLaVA-NeXT series regarding data scaling, model configuration, and visual feature representations.
In our evaluation, it is used in the spatial VideoQA table to test a unified image/video MLLM under quality variation.

\noindent \textbf{MiniCPM-O 2.6~\cite{openbmb2025minicpmo26}}:
MiniCPM-O 2.6 is a compact open-source multimodal model in the MiniCPM-O series.
The technical report presents it as an MLLM for vision, speech, and multimodal live streaming on resource-constrained devices.
We include it in the spatial VideoQA table as a compact open-source baseline with multimodal streaming capability.

\noindent \textbf{MiniCPM-V 4.5~\cite{minicpmv2025_45}}:
MiniCPM-V 4.5 is an 8-billion-parameter vision-language model built on a Qwen3-8B language backbone and a SigLIP2-400M vision encoder.
The technical report presents it as a model for single-image, multi-image, and high-FPS video understanding on mobile hardware.
We evaluate it in the spatial VideoQA table to compare efficient mobile-oriented MLLMs with larger open-source baselines.

\noindent \textbf{Phi-4-multimodal~\cite{microsoft2025phi4mm}}:
Phi-4-multimodal is Microsoft's fully multimodal Phi-4 model that accepts text, image, and audio inputs and generates text outputs.
It builds on the research and datasets used for the Phi-3.5 and Phi-4 language models, extending them to support rich visual and speech grounding.
We include it in the spatial VideoQA table as a compact open-source baseline for multimodal reasoning under degraded and restored video inputs.

\noindent \textbf{Qwen3-VL-8B~\cite{qwen2025qwen3vl}}:
Qwen3-VL-8B is an 8-billion-parameter vision-language model in the Qwen3-VL family.
It offers upgraded text generation, deeper visual perception and reasoning, and extended context length compared with earlier Qwen-VL releases.
The model is evaluated in the spatial VideoQA table as a competitive zero-shot open-source baseline.

\noindent \textbf{Qwen2.5-VL-7B~\cite{qwen2025v25vl}}:
Qwen2.5-VL-7B is a 7-billion-parameter member of the Qwen2.5-VL series, pretrained on trillions of tokens for vision-language understanding.
It incorporates architectural refinements such as window attention in the vision encoder and dynamic frame-rate sampling to better support video across diverse temporal resolutions.
The model is included in both final VideoQA tables and also serves as the base model for our ClearText-Video supervised fine-tuning experiment.

\noindent \textbf{Qwen2.5-VL-7B-SFT}:
Qwen2.5-VL-7B-SFT is our instruction-tuned variant of Qwen2.5-VL-7B trained on the ClearText-Video training set.
It is reported in the spatial VideoQA table to evaluate whether dataset-specific supervision improves robustness across HQ, degraded, and restored inputs.
We do not include this tuned variant in the temporal quantitative table, where the comparison focuses on zero-shot proprietary and open-source MLLMs.

\noindent \textbf{VideoLLaMA3-7B~\cite{zhang2025videollama}}:
VideoLLaMA3-7B is a 7-billion-parameter multimodal foundation model from the VideoLLaMA 3 series that targets high-quality image and video understanding.
The accompanying project reports strong performance among 7B-sized models on video benchmarks such as LVBench and VideoMME.
VideoLLaMA3-7B is included in the temporal VideoQA table as a video-specialized open-source baseline for reasoning over text trajectories and visibility changes.

\noindent \textbf{Kimi-VL-16B~\cite{team2025kimi}}:
Kimi-VL-16B is an efficient 16-billion-parameter Mixture-of-Experts vision-language model developed by Moonshot AI.
Although the total parameter count is 16B, only about 2.8B parameters are activated during inference, reducing active computation relative to the model's total parameter count.
The technical report highlights long-context and high-resolution visual understanding, and we include Kimi-VL-16B in the temporal VideoQA table as the highest-scoring open-source baseline under most temporal quality conditions.

\subsection{Additional Diagnostic Analyses}
\label{sec:additional_diagnostics}

\begin{figure}[t]
    \centering
    \includegraphics[width=\textwidth]{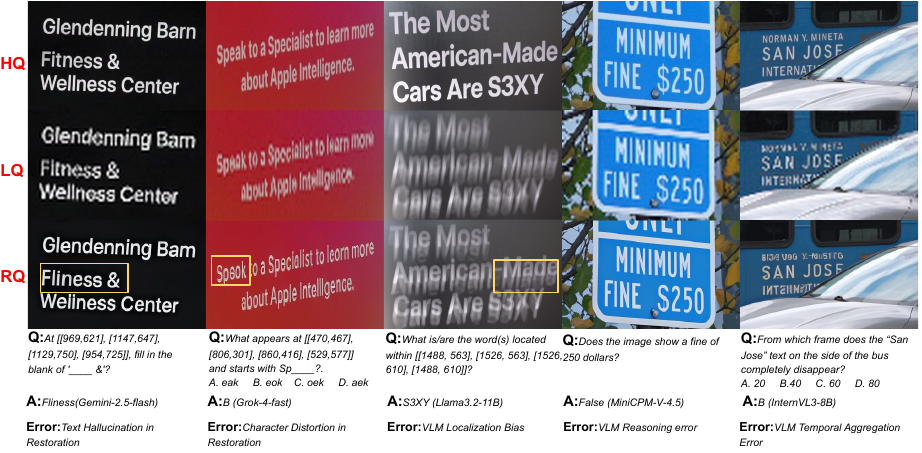}
    \caption{Failure modes on RQ inputs. Restoration may hallucinate or distort text, while MLLMs may further fail through localization bias, reasoning errors, or temporal aggregation errors.}
    \label{fig:rq_failure}
\end{figure}

We provide three additional diagnostic analyses. First, Fig.~\ref{fig:rq_failure} groups RQ failures into restoration-side artifacts and MLLM-side errors. Restoration models can introduce spurious or distorted character strokes before reasoning begins, while MLLMs may still fail due to localization bias, semantic reasoning errors, or temporal aggregation mistakes. This helps explain why visually sharper restored videos do not always improve downstream QA.

Second, Table~\ref{tab:degradation_analysis} isolates low-resolution and blur degradations on the same videos and QA pairs. Both reduce spatial QA accuracy, but blur is more harmful: averaged over 16 MLLMs, LR reduces accuracy by 3.14 points from HQ, whereas blur reduces it by 6.05 points. The same trend appears for Qwen3-VL-8B, which is consistent with greater sensitivity of character-level reasoning to corrupted strokes and text boundaries.

\begin{table}[t]
\centering
\caption{Quality-type comparison on spatial QA. Avg. Acc is averaged over the 16 MLLMs in the main spatial VideoQA table.}
\label{tab:degradation_analysis}
\setlength{\tabcolsep}{5pt}
\begin{tabular}{l|cccccc}
\toprule
 & HQ & LR & Blur & DOVE & MIMO & S3Diff \\
\midrule
Avg. Acc$\uparrow$      & 47.73 & 44.59 & 41.69 & 45.21 & 44.02 & 44.79 \\
Avg. $\Delta$HQ         & --    & -3.14 & -6.05 & -2.52 & -3.72 & -2.95 \\
Qwen3-VL-8B Acc$\uparrow$     & 50.56 & 45.43 & 44.46 & 47.70 & 46.91 & 46.64 \\
Qwen3-VL-8B $\Delta$HQ        & --    & -5.13 & -6.10 & -2.86 & -3.65 & -3.92 \\
\bottomrule
\end{tabular}
\end{table}

Third, Table~\ref{tab:ocr_llm} adds an OCR-mediated text-only baseline. PaddleOCR~\cite{paddleocr} extracts text from each quality version, and Qwen2.5-7B answers using only the OCR text without image/video input. OCR+LLM remains far below direct Qwen2.5-VL-7B, indicating that the OCR-only pipeline is insufficient for this benchmark and suggesting that spatial grounding and multimodal reasoning are important. RQ improves Qwen2.5-VL-7B over its DQ results but does not improve OCR+LLM, suggesting that restoration may benefit direct visual reasoning while still producing outputs that remain difficult for OCR or contain text distortions. HQ denotes OCR from the original HQ inputs, not oracle human transcripts.

\begin{table}[t]
\centering
\caption{OCR+LLM vs. direct MLLM on spatial QA. OCR+Qwen2.5-7B uses OCR text only, without image/video input.}
\label{tab:ocr_llm}
\footnotesize
\setlength{\tabcolsep}{4.5pt}
\begin{tabular}{@{}l|ccc|ccc@{}}
\toprule
\multirow{2}{*}{Input quality} 
& \multicolumn{3}{c|}{OCR+Qwen2.5-7B} 
& \multicolumn{3}{c}{Qwen2.5-VL-7B} \\
\cmidrule(lr){2-4}\cmidrule(lr){5-7}
& Acc$\uparrow$ & UAcc$\uparrow$ & OC$\downarrow$ 
& Acc$\uparrow$ & UAcc$\uparrow$ & OC$\downarrow$ \\
\midrule
HQ                 & \textbf{23.3} & 66.0 & 14.6 & 51.21 & 58.69 & 10.89 \\
DQ-Low\_res        & 22.7 & \textbf{66.7} & \textbf{13.5} & 40.23 & 61.67 & 3.99 \\
DQ-Blur            & 22.3 & 66.6 & 13.7 & 41.72 & 62.66 & 9.19 \\
RQ-DOVE            & 22.9 & 64.5 & 17.7 & 46.58 & 59.97 & 10.79 \\
RQ-MIMO            & 22.2 & 61.9 & 20.1 & 46.10 & 59.89 & 10.79 \\
RQ-S3DIFF          & 21.4 & 64.7 & 16.7 & 45.66 & 60.74 & 10.14 \\
\bottomrule
\end{tabular}
\end{table}

\begin{figure*}[t]
    \centering
    \includegraphics[width=1.0\linewidth]{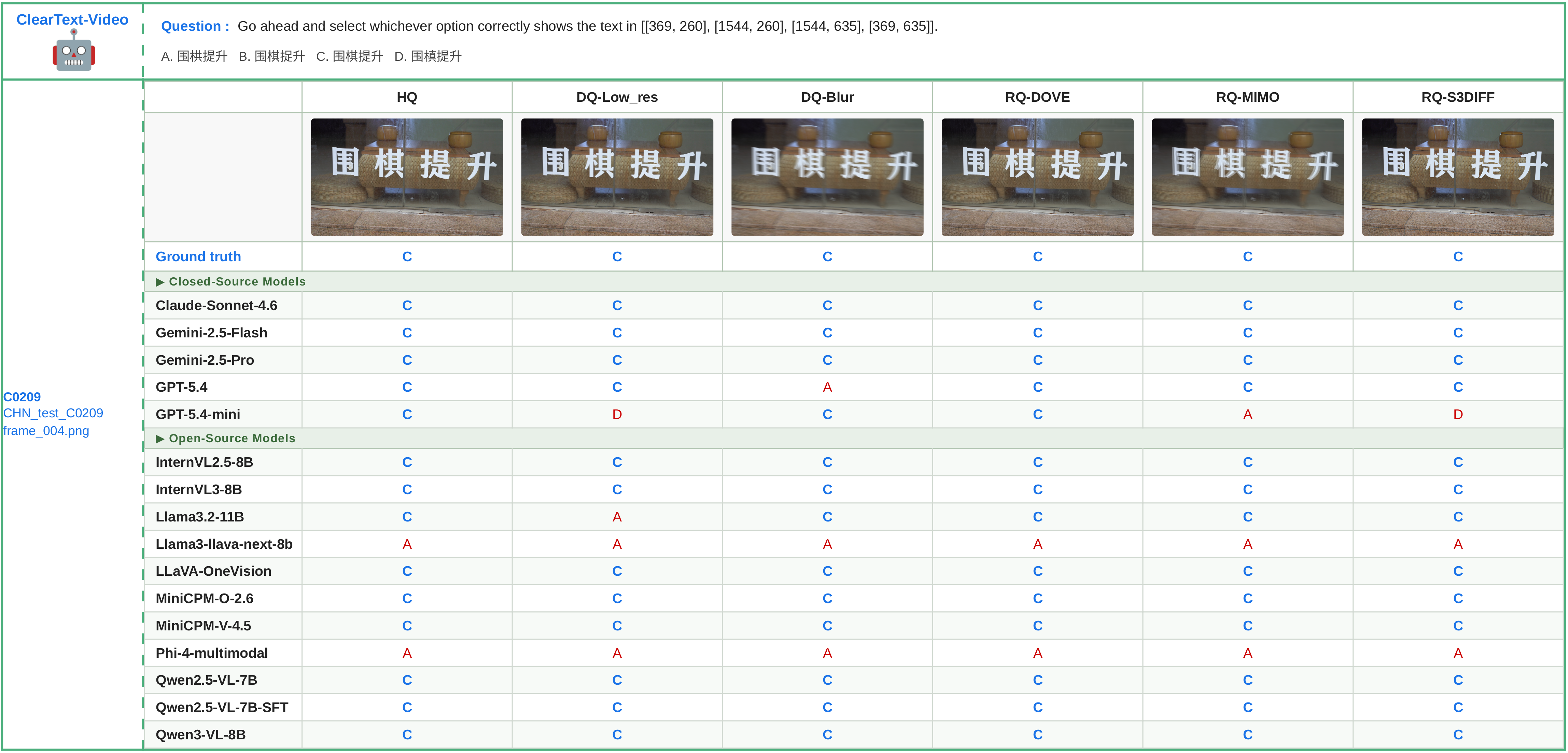}
    \caption{\textbf{Result visualization on ClearText-Video for text-centric spatial VideoQA.}
    This example shows a Chinese sign in which the question asks the model to select the option that correctly matches the text inside a specified bounding box. Most higher-performing models identify the correct phrase consistently across quality conditions, while weaker models confuse visually similar Chinese characters or repeatedly select an incorrect option. This example illustrates that even when the text is visually prominent, robust spatial text understanding still depends on precise character-level discrimination rather than coarse recognition.}
    \label{fig:vis_spatial_single5}
\end{figure*}

\begin{figure*}[t]
    \centering
    \includegraphics[width=1.0\linewidth]{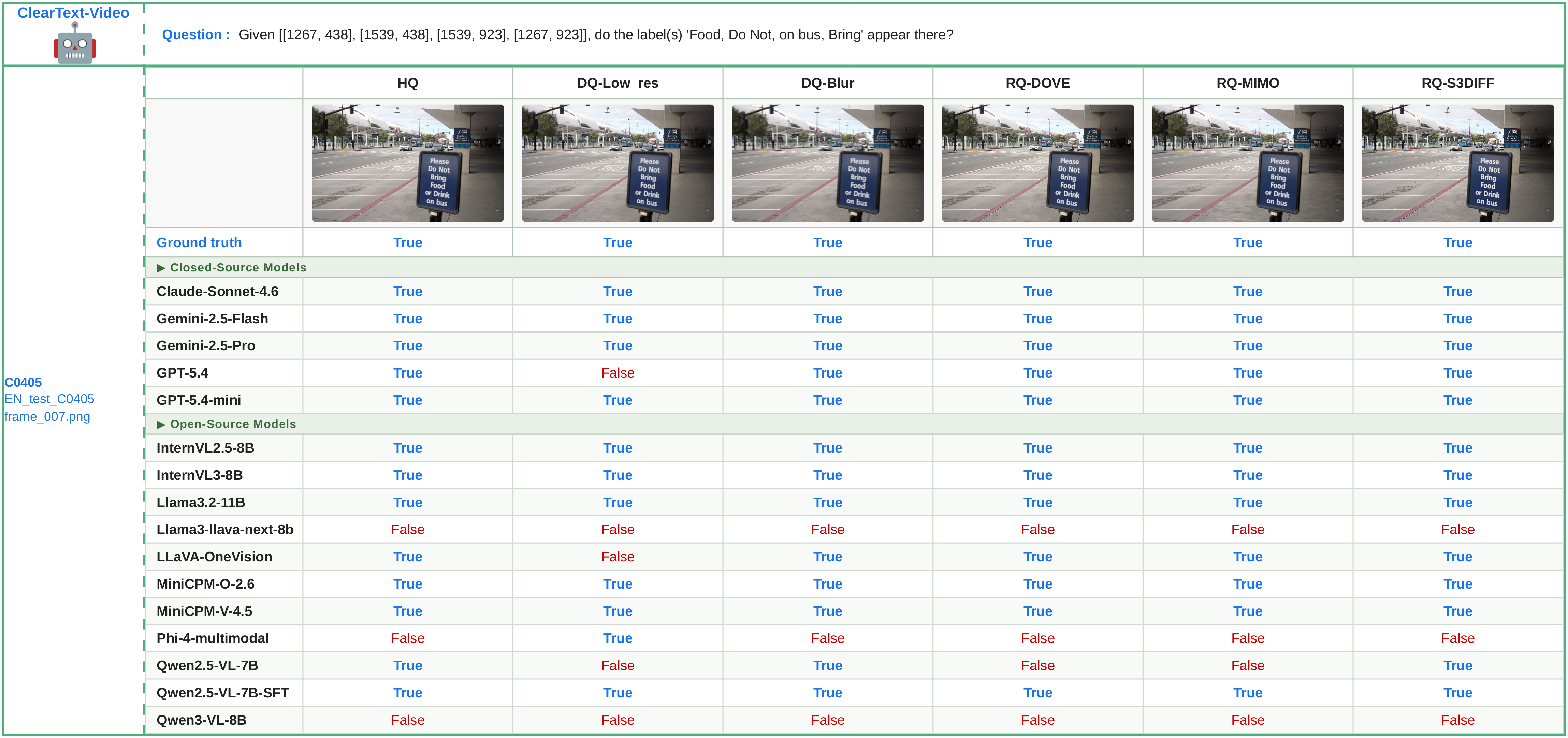}
    \caption{\textbf{Result visualization on ClearText-Video for text-centric spatial VideoQA.}
    This example shows an outdoor transit sign in which the question asks whether a set of words, including ``Food'', ``Do Not'', ``on bus'', and ``Bring'', appears inside the specified bounding box. Many models correctly verify the presence of the target phrases across quality conditions, but some models consistently reject the correct answer or become unstable under degraded and restored inputs. This case illustrates that spatial VideoQA requires not only text recognition, but also accurate grounding of multiple words within the queried region.}
    \label{fig:vis_spatial_single6}
\end{figure*}

\begin{figure*}[t]
    \centering
    \includegraphics[width=1.0\linewidth]{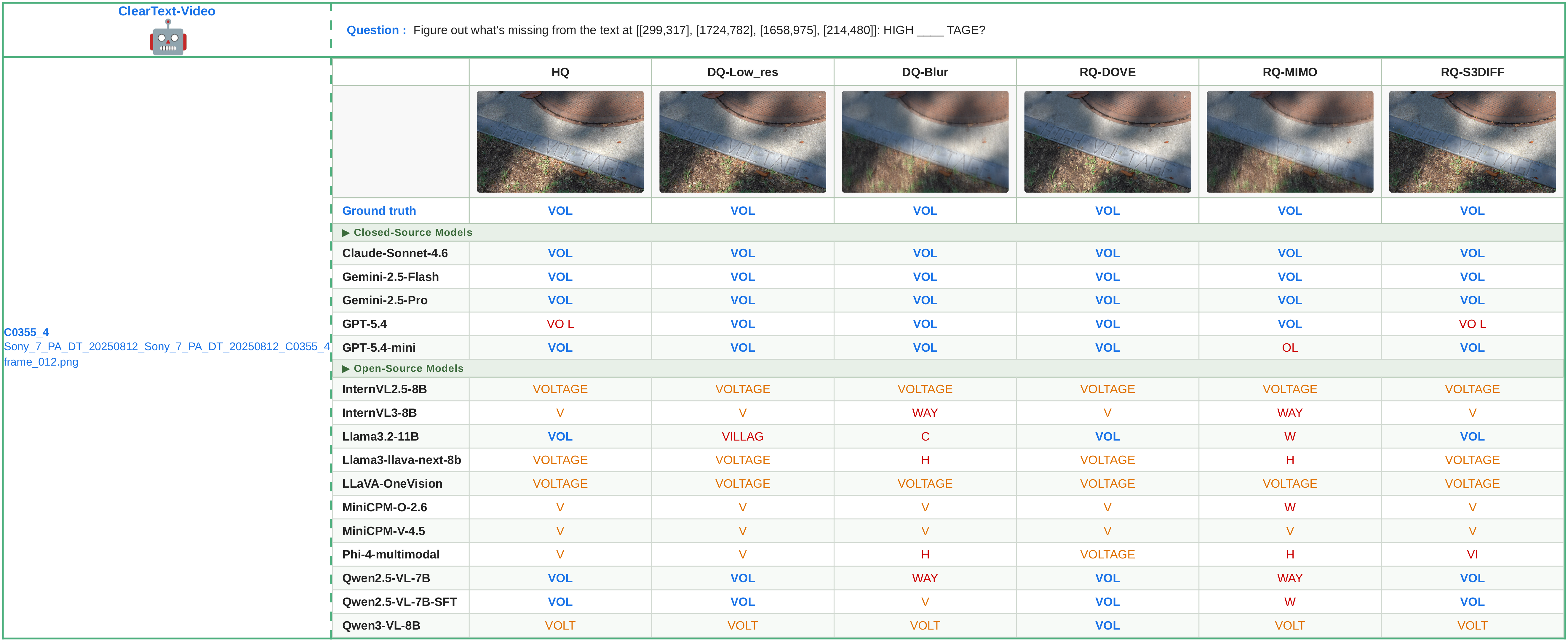}
    \caption{\textbf{Result visualization on ClearText-Video for text-centric spatial VideoQA.}
    This example shows an English warning sign where the question requires completing the missing substring in ``HIGH \_\_\_\_ TAGE''. Most proprietary MLLMs consistently recover the correct missing text across different quality conditions, while several open-source models either output the full word instead of the missing span, predict only a single character, or hallucinate visually plausible but incorrect fragments. The results highlight the difficulty of fine-grained text completion under blur, low resolution, and restoration artifacts.}
    \label{fig:vis_spatial_single9}
\end{figure*}

\subsection{Visualization for Text-Centric Video Restoration}

\begin{figure*}[htb]
    \centering
    \caption{Qualitative results on the Text-Centric Video Restoration benchmark on the CTVid test set. Results are shown for both English and Chinese text instances. The examples include video super-resolution and deblurring outputs, highlighting differences in text fidelity and legibility across methods.}
    \label{fig:appendix_Vis_results}
    \begin{subfigure}[htb]{0.91\textwidth}
        \centering
        \includegraphics[width=0.91\linewidth]
        {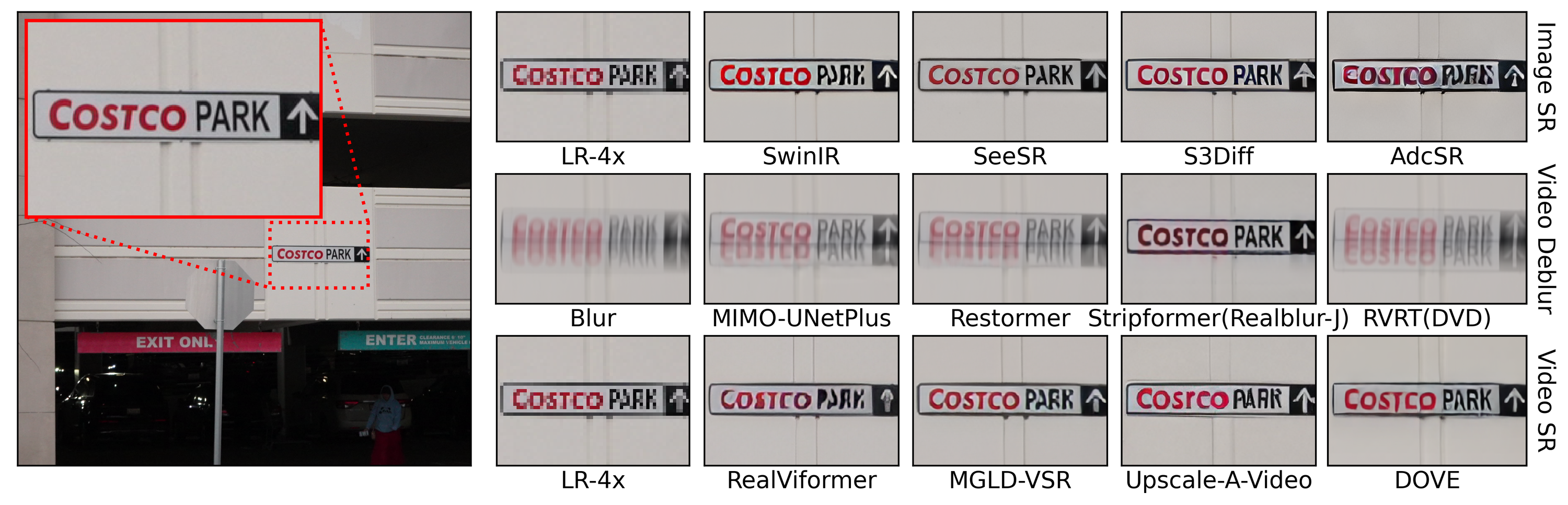}
        \caption{A parking lot entrance.}
        \label{fig:Vis_results_a}
    \end{subfigure}

    \begin{subfigure}[htb]{0.91\textwidth}
        \centering
        \includegraphics[width=0.91\linewidth]
        {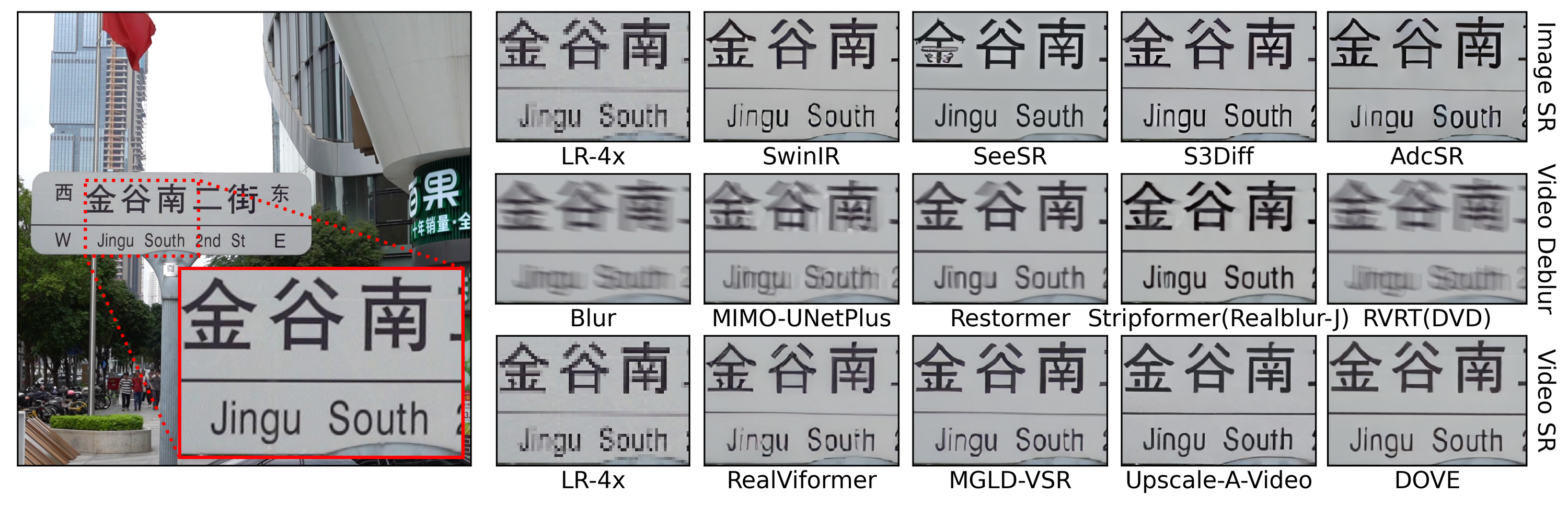}
        \caption{A bilingual street sign.}
        \label{fig:Vis_results_b}
    \end{subfigure}
    
    \begin{subfigure}[htb]{0.91\textwidth}
        \centering
        \includegraphics[width=0.91\linewidth]{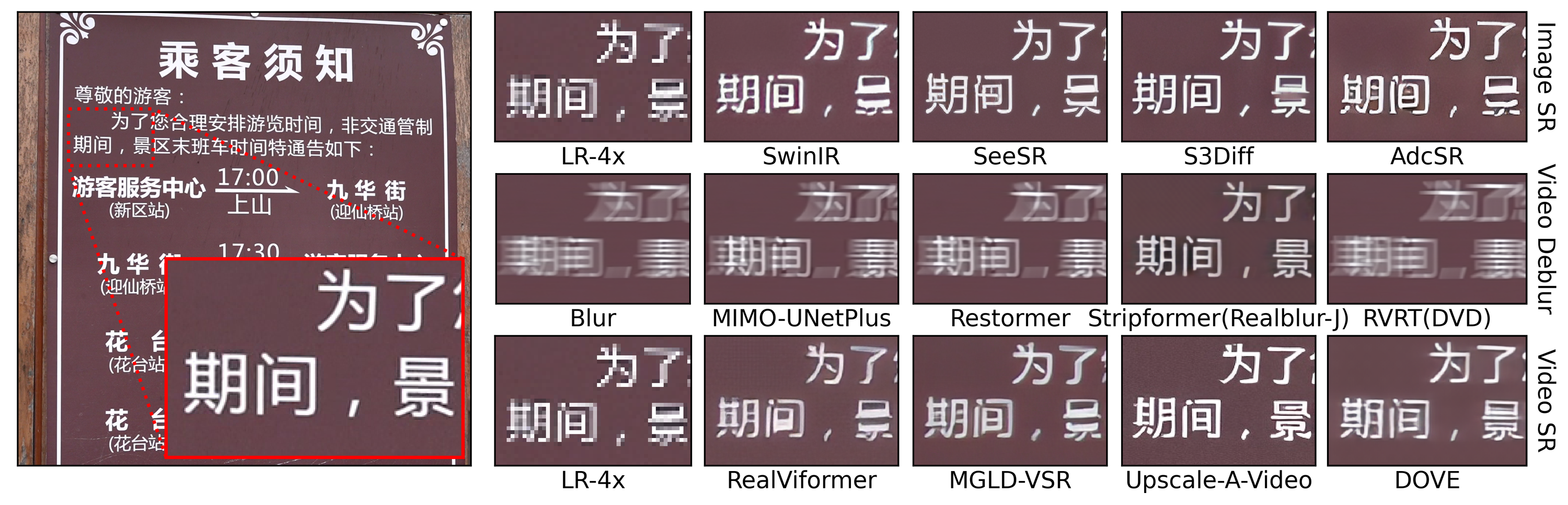}
        \caption{A bus schedule sign.}
        \label{fig:Vis_results_c}
    \end{subfigure}
    
    \begin{subfigure}[htb]{0.91\textwidth}
        \centering
        \includegraphics[width=0.91\linewidth]
        {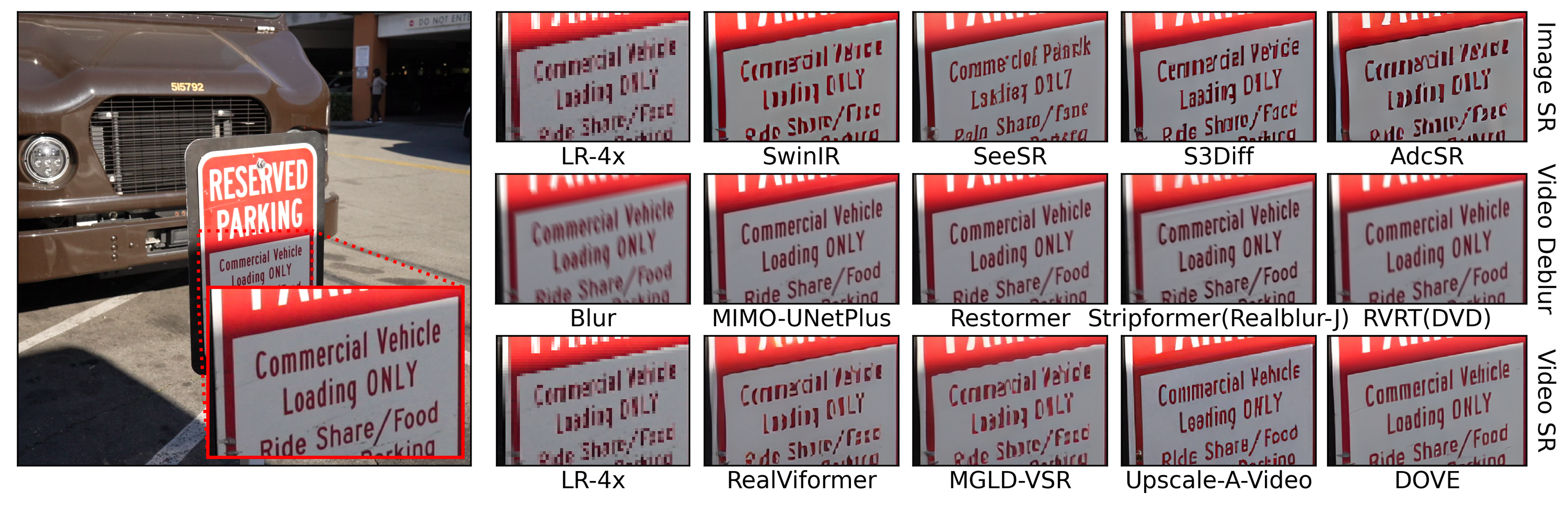}
        \caption{A ``Reserved Parking'' sign.}
        \label{fig:Vis_results_d}
    \end{subfigure}
\end{figure*}
We present additional examples from CTVid and evaluate several video restoration algorithms on these cases. The results illustrate the strengths and weaknesses of each algorithm across English and Chinese text. 

\noindent \textbf{Video deblurring}
Stripformer produces relatively clear outputs in these examples, but it substantially changes the color of the ``COSTCO'' sign in Figure~\ref{fig:Vis_results_a} and changes ``Jingu'' to ``Jimgu'' in Figure~\ref{fig:Vis_results_b}. MIMO-UNet+, Restormer, and RVRT provide only limited deblurring in Figures~\ref{fig:Vis_results_a}, \ref{fig:Vis_results_b}, and \ref{fig:Vis_results_c}.

\noindent \textbf{Image super-resolution} The tested approaches perform well on larger English text, such as ``Jingu South'' in Figure~\ref{fig:Vis_results_b}. However, SeeSR renders ``South'' as ``Sauth''. In the shown examples, all evaluated methods struggle with the dense details of Chinese characters in Figure~\ref{fig:Vis_results_c} and the smaller text in Figure~\ref{fig:Vis_results_d}. Several methods generate illegible or unrelated text; for example, SeeSR turns ``Ride Share'' into ``Paln Sharo''.

\noindent \textbf{Video super-resolution} Video super-resolution methods appear to preserve text more accurately in these examples. Upscale-A-Video recovers entire words but occasionally changes individual letters (\textit{e.g.}, ``Commarcial'' instead of ``Commercial'' and ``$L\theta$ading'' instead of ``Loading''). DOVE slightly deforms ``COSTCO'' in Figure~\ref{fig:Vis_results_a} and omits a few strokes in Figure~\ref{fig:Vis_results_c}.

Existing methods can introduce text distortions beyond those in the degraded input, potentially harming downstream performance. In these examples, video-based methods often preserve text more accurately than image-only methods by exploiting temporal information, but they still struggle with compact Chinese characters. These results motivate greater use of temporal priors in future diffusion-based video restoration research.

\subsection{Spatial VideoQA Visualization under Multi-Quality Videos}

As shown in Figs.~\ref{fig:vis_spatial_single5},~\ref{fig:vis_spatial_single6},~\ref{fig:vis_spatial_single9},
we provide representative qualitative visualizations for text-centric spatial VideoQA on ClearText-Video under multiple video-quality conditions.
Each example corresponds to a dynamic video clip with a localized scene-text region and a spatially grounded question, such as reading full street names, identifying text on outdoor signs, or completing missing characters within bounding boxes.
For each scene, we construct one spatial question grounded in a specific text group and render the corresponding clip under six input-quality conditions: HQ, DQ-Low\_res, DQ-Blur, RQ-DOVE, RQ-MIMO, and RQ-S3DIFF.
For every quality condition, we show a representative keyframe for context together with the question, the ground-truth answer, and predictions from 16 MLLMs.

These qualitative results clarify the quantitative trends in Table~\ref{tab:all-imageQA_results} and highlight the fine-grained difficulty of spatial text understanding in realistic videos.
Although many models can roughly recognize the target text, they often fail on details such as missing tokens, incorrect ordering, visually similar answer options, or incomplete answers to fill-in-the-blank questions.
When moving from HQ to degraded videos, several models show weaker spatial grounding and appear to rely excessively on partial OCR cues.
Moreover, restored videos do not always lead to better spatial reasoning: even when the text appears perceptually sharper, several models still output incorrect answers, suggesting that current restoration methods may improve visual appearance without fully recovering text semantics.
Our instruction-tuned Qwen2.5-VL-7B-SFT exhibits more stable predictions across quality conditions, consistent with its greater robustness in the numerical evaluation.
Overall, these visualizations illustrate challenging spatial text-reasoning cases in ClearText-Video and reveal failure patterns that are not evident from aggregate metrics alone.
More detailed, example-specific analyses of model behavior are provided in the captions of Figs.~\ref{fig:vis_spatial_single5}, \ref{fig:vis_spatial_single6}, and \ref{fig:vis_spatial_single9}.

\subsection{Temporal VideoQA Visualization under Multi-Quality Videos}

\begin{figure*}[t]
    \centering
    \includegraphics[width=1.0\linewidth]{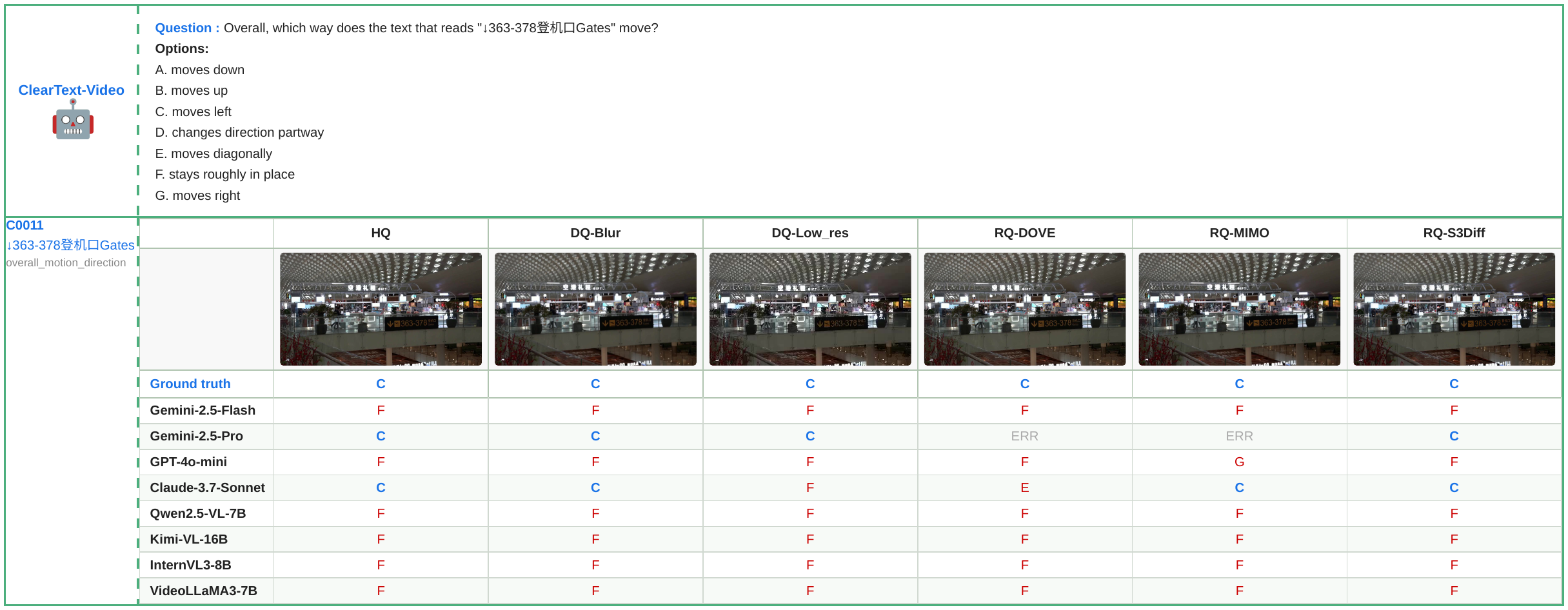}
    \caption{\textbf{Result visualization on ClearText-Video for text-centric temporal VideoQA.}
    This example asks for the overall motion direction of the text ``$\downarrow$363-378Gates'' across the video clip. Although the ground-truth motion is consistently leftward under all quality conditions, most models incorrectly predict that the text stays roughly in place, suggesting substantial reliance on static keyframe cues in this example. Only a few models predict the correct motion direction, while their predictions still become unstable under degraded or restored inputs, showing that robust temporal motion reasoning over scene text remains challenging for current MLLMs.}
    \label{fig:vis_temporal_single1}
\end{figure*}

\begin{figure*}[t]
    \centering
    \includegraphics[width=1.0\linewidth]{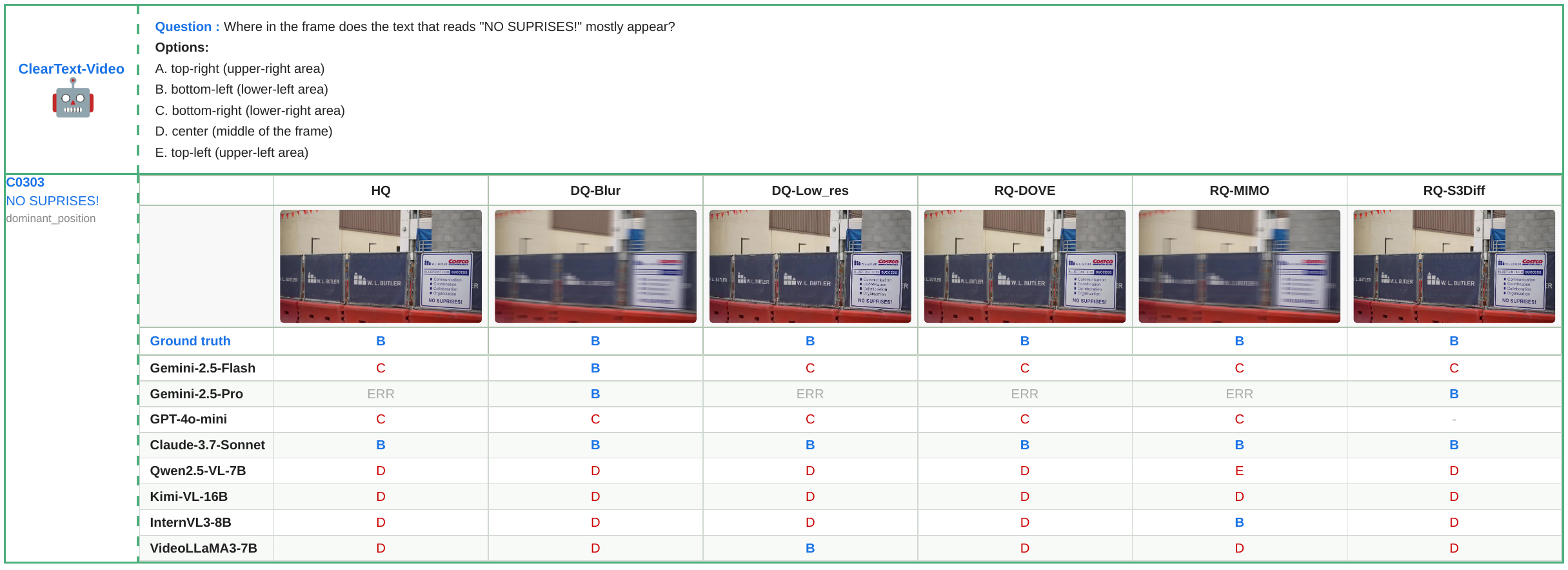}
    \caption{\textbf{Result visualization on ClearText-Video for text-centric temporal VideoQA.}
    This example asks where the text ``NO SUPRISES!'' appears for most of the clip. The ground-truth answer is consistently the bottom-left region, but many models confuse it with the bottom-right or center regions, and some predictions vary across different quality conditions. This example illustrates that reliably estimating the dominant spatial position of a text instance over time requires more than recognizing the text in a single frame, as models must aggregate its location across the full temporal sequence.}
    \label{fig:vis_temporal_single2}
\end{figure*}

\begin{figure*}[t]
    \centering
    \includegraphics[width=1.0\linewidth]{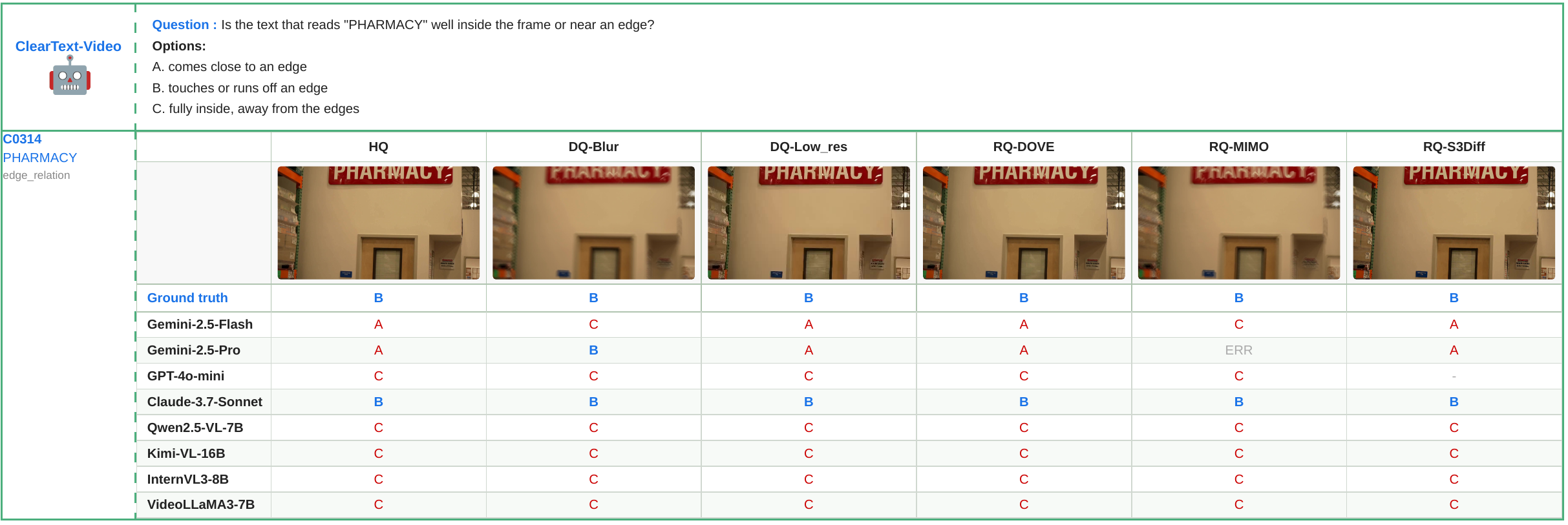}
    \caption{\textbf{Result visualization on ClearText-Video for text-centric temporal VideoQA.}
    This example asks about the relation between the text ``PHARMACY'' and the frame boundary. Although the correct answer is that the text touches or runs off an edge, most models incorrectly classify it as fully inside or merely close to an edge, especially under degraded and restored inputs. This case highlights the difficulty of boundary-aware temporal text reasoning, where models must precisely track the relation between a text instance and the frame boundary rather than relying on coarse recognition alone.}
    \label{fig:vis_temporal_single3}
\end{figure*}

As shown in Figs.~\ref{fig:vis_temporal_single1},~\ref{fig:vis_temporal_single2},~\ref{fig:vis_temporal_single3}, 
we provide qualitative visualizations for text-centric temporal VideoQA on ClearText-Video under different video-quality conditions.
Each example is a 120-frame clip in which the question requires reasoning about the temporal behavior of a specific text instance, such as whether its width ever exceeds half of the screen or in how many frames a target phrase remains visible.
For each dynamic scene, we consider the same clip under five quality conditions: HQ, DQ-Blur, RQ-DOVE, RQ-MIMO, and RQ-S3DIFF. We then pose a single temporal question about the evolution of the target text instance over time.
For every quality condition, we display a representative keyframe together with the question, the ground-truth answer, and predictions from the selected set of eight temporal models.
Although the visualization shows only a single frame for brevity, the answer always depends on the full temporal sequence.
These qualitative results complement the accuracy trends in Table~\ref{tab:vqa_temporal_results} and reveal that temporal decisions are sensitive to blur and restoration artifacts. Even for the same motion trajectory, models can produce different estimates of visibility duration or threshold crossing across video-quality conditions, causing inconsistent answers.
The examples further show that many models appear to rely on static cues and fail to aggregate temporal evidence correctly, resulting in substantial underestimation or overestimation of visibility duration and incorrect temporal decisions.
Models that perform well on spatial VideoQA may still struggle on these temporal questions, indicating that temporal text reasoning is not merely a byproduct of strong OCR, but requires effective aggregation of visibility, motion, and duration cues over time.
While proprietary models are comparatively more robust in aggregate, the qualitative examples show that even strong models still exhibit noticeable fluctuations across quality conditions.
Overall, these visualizations show that ClearText-Video provides a challenging, quality-aware benchmark for temporal text-centric reasoning and exposes failure patterns that are not evident from aggregate metrics alone.
More detailed, example-level analyses are provided in the captions of Figs.~\ref{fig:vis_temporal_single1}--\ref{fig:vis_temporal_single3}.


\begin{table}[t]
\centering
\small
\renewcommand{\arraystretch}{1.15}
\caption{System prompt used for evaluating spatial video question answering.}
\label{tab:system_prompt_video_spatial}

\begin{tabular}{p{0.96\linewidth}}
\hline
\centering \textbf{System Prompt} \tabularnewline
\hline

\textbf{Fill-in Question Prompt} \\

Your task is to answer fill-in questions based on image input.

\textbf{Input Format:}
An image with \{size\} and a fill-in question string, e.g., ``C\_\_\_\_T''.

\textbf{Output Format:}
A single-line JSON object:
\texttt{\{"answer": "<only the missing characters>", "reasoning": "<a brief explanation, 1 sentence max>"\}}

\textbf{Rules:}
Only output the missing characters, not the full word. Keep the reasoning short and image-based. Do not infer from common sense or spelling patterns, because the answer may not follow normal spelling. Only answer based on what is actually visible in the image. The blank, represented as ``\_\_\_\_'', can correspond to any number or type of characters, including letters, digits, Chinese characters, or symbols.

\textbf{Example:}
Question: What letter is missing in ``C\_\_T''?  
Answer: \texttt{\{"answer": "a", "reasoning": "C-a-T forms CaT as shown in the image."\}}

Respond only with a JSON object. Do not include anything else.

\\
\hline

\textbf{Multi-choice Question Prompt} \\

You are a vision-language assistant specialized in text recognition. Your task is to answer multiple-choice questions based on image content.

\textbf{Input Format:}
An image with \{size\} and a question string with options, e.g., ``Which letter is missing in G\_\_FT? Choices: A. A B. B C. I D. H''.

\textbf{Output Format:}
A single-line JSON object:
\texttt{\{"answer": "<only the correct letter(s)>", "reasoning": "<a brief explanation, 1 sentence max>"\}}

\textbf{Rules:}
Only output the letter(s) corresponding to the correct option. Do not repeat the full option. Keep the reasoning short and image-based. The answer can contain multiple letters, e.g., ``A,C'', if multiple options are correct. Do not infer from common sense or spelling patterns. Only answer based on what is actually visible in the image.

\textbf{Example:}
Question: Which word(s) is located within the area [\ldots]? Choose the correct option.  
A. Apple \quad B. App \quad C. Able \quad D. Aple  

Answer: \texttt{\{"answer": "A,B", "reasoning": "I can see both `Apple' and `App' in the bounding box area."\}}

Respond only with a JSON object. Do not include anything else.

\\
\hline

\textbf{True-False Question Prompt} \\

You are a vision-language assistant specialized in text recognition. Your task is to answer true/false questions based on the content of the image.

\textbf{Input Format:}
An image with \{size\} and a true/false question string, e.g., ``The word in the image is DOG. True or False?''.

\textbf{Output Format:}
A single-line JSON object:
\texttt{\{"answer": "True" or "False", "reasoning": "<a brief explanation, 1 sentence max>"\}}

\textbf{Rules:}
The answer must be strictly ``True'' or ``False''. Keep the reasoning short and image-based. Do not infer from common sense or spelling patterns. Only answer based on what is actually visible in the image.

\textbf{Example:}
Question: The word in the image is DOG. True or False?  
Answer: \texttt{\{"answer": "False", "reasoning": "The image shows DoG, not DOG."\}}

Respond only with a JSON object. Do not include anything else.

\\
\hline
\end{tabular}
\end{table}

\begin{table}[t]
\centering
\footnotesize
\renewcommand{\arraystretch}{1.15}
\caption{System prompt used for evaluating temporal video question answering.}
\label{tab:system_prompt_video_temporal}

\begin{tabular}{p{0.96\linewidth}}
\hline
\centering \textbf{System Prompt} \tabularnewline
\hline

You are a helpful assistant that can answer questions about a video.

\textbf{Video context:}
The original video has \{total\_frames\} frames in total\{fps\_clause\}.
Original frame resolution: \{original\_size\}.
Frames are resized to \{resized\_size\} before being shown to you.
Any coordinates or bounding boxes mentioned in the questions are based on the original resolution \{\linebreak original\_size\}.
You are shown \{n\_sampled\} frames uniformly sampled from this video.
The frames are given in chronological order.
The k-th frame you see, 1-based, corresponds to original frame index sampled\_indices[k-1], where sampled\_indices = \{sampled\_indices\_str\}.
When the question refers to a specific frame index or timestamp, reason about which of your sampled frames is closest to it.
You only have access to the listed sampled frames.

\textbf{Answer vocabulary --- geometric question types:}
Each question about a specific target text uses one of the fixed option sets below.
Pick the option that best matches what you observe across the sampled frames.

\textbf{Position regions:}
``center (middle of the frame)'': the central $\sim$1/3 of the frame on both the horizontal and vertical axes;
``top-left (upper-left area)'': upper half of the frame, left of centre;
``top-right (upper-right area)'': upper half of the frame, right of centre;
``bottom-left (lower-left area)'': lower half of the frame, left of centre;
``bottom-right (lower-right area)'': lower half of the frame, right of centre.

\textbf{Position change amount:}
``stays in roughly the same place (barely shifts)'': the text barely moves; displacement is negligible;
``moves a moderate amount (shifts across part of the frame)'': noticeable shift --- text crosses a portion of the frame;
``moves a large distance (shifts across much of the frame)'': large displacement --- text traverses most of the frame.

\textbf{Overall motion direction:}
``stays roughly in place'': negligible net displacement throughout;
``moves left'': predominantly leftward drift;
``moves right'': predominantly rightward drift;
``moves up'': predominantly upward drift;
``moves down'': predominantly downward drift;
``moves diagonally'': significant movement on both horizontal and vertical axes simultaneously;
``changes direction partway'': reverses along its dominant axis mid-clip, e.g., moves right then left.

\textbf{Text size level:}
``small (a minor part of the frame)'': text occupies a small fraction of the frame, roughly $<2\%$;
``medium (a noticeable block of the frame)'': clearly visible, moderate fraction, roughly $2$--$6\%$;
``large (a major part of the frame)'': covers a substantial portion of the frame, roughly $6$--$15\%$;
``very large (dominates the frame)'': fills most of the frame, roughly $>15\%$.

\textbf{Scale change:}
``stays about the same size'': size is roughly constant throughout the clip;
``gets larger'': text grows noticeably from the start to the end;
``gets smaller'': text shrinks noticeably from the start to the end;
``size fluctuates (goes up and down)'': size oscillates non-monotonically during the clip.

\textbf{Edge relation:}
``fully inside, away from the edges'': text stays well clear of all frame borders throughout;
``comes close to an edge'': text approaches a border but does not touch it;
``touches or runs off an edge'': text reaches or extends beyond a frame border.

Answer the question based on the video.
Directly return the answer with no extra text.

\\
\hline
\end{tabular}
\end{table}


\clearpage

\end{document}